\documentclass[10pt,journal,compsoc]{IEEEtran} %
\usepackage[english]{babel}
\usepackage{cite}
\usepackage{bm}
\usepackage{amssymb,amsfonts}
\usepackage{algorithmic}
\usepackage{graphicx}
\usepackage{textcomp}
\usepackage{booktabs}
\usepackage{array, caption, threeparttable}
\usepackage{wrapfig}

\usepackage[cmex10]{amsmath}

\usepackage[caption=false,font=footnotesize,labelfont=sf,textfont=sf]{subfig}

\usepackage{booktabs}
\usepackage{multirow}

\usepackage{color}
\usepackage{pifont}%

\usepackage{url}

\usepackage{siunitx}
\DeclareSIUnit{\nothing}{\relax}

\usepackage{cite} %

\newcommand\MYhyperrefoptions{bookmarks=true,bookmarksnumbered=true,
pdfpagemode={UseOutlines},plainpages=false,pdfpagelabels=true,
colorlinks=true,citecolor={black},
urlcolor={black},
pdftitle={Copy-Move Image Forgery Detection Based on Evolving Circular Domains Coverage},%
pdfsubject={Computer Vision},%
pdfauthor={Shilin Lu, Xinghong Hu, Chengyou Wang, Lu Chen, Shulu Han, Yuejia Han},%
pdfkeywords={Image forensics, copy-move forgery detection (CMFD), scale invariant feature transform (SIFT), speed-up robust feature (SURF), evolving circular domains coverage (ECDC)}}%
\usepackage[\MYhyperrefoptions,pdftex,hidelinks]{hyperref}

\usepackage{adjustbox} %

\usepackage{etoolbox} %

\newif\ifclearsectionlook
\clearsectionlookfalse

\newif\ifclearsubseclook
\clearsubseclookfalse

\newif\iflongversion
\longversiontrue

\usepackage{xcolor}

\newif\ifblacktext
\blacktexttrue

\ifblacktext
\else
\fi

\usepackage[draft,authormarkup=none]{changes}
\definechangesauthor[color=blue]{GG}
\definechangesauthor[name=TD, color=magenta]{TD}
\definechangesauthor[name=GO, color=orange]{GO}
\definechangesauthor[name=CB, color=violet]{CB}

\definecolor{somegray}{rgb}{0.5, 0.5, 0.5}
\newcommand{\darkgrayed}[1]{\textcolor{somegray}{#1}}
\makeatletter
\newcommand*\titleheader[1]{\gdef\@titleheader{#1}}
\AtBeginDocument{%
  \let\st@red@title\@title
  \def\@title{%
    \vskip-1.3em
    \bgroup\normalfont\large\centering\@titleheader\par\egroup
    \vskip1.5em\st@red@title}
}
\makeatother

\titleheader{\darkgrayed{\\}}

\title{Copy-Move Image Forgery Detection Based on Evolving Circular Domains Coverage}

\begin{document}

\author{Shilin Lu,
Xinghong Hu,
Chengyou Wang,
Lu Chen,
Shulu Han,
Yuejia Han

\thanks{This work was supported in part by the Shandong Provincial Natural Science Foundation, China (Nos. ZR2021MF060, ZR2017MF020), in part by the  Education and Teaching Reform Research Project of Shandong University, Weihai (No. Y2021054), in part by the National Natural Science Foundation of China (No. 61702303), in part by the Science and Technology Development Plan Project of Weihai Municipality in 2020, and in part by the 14th Student Research Training Program (SRTP) at Shandong University, Weihai (No. A19167).
}%
\thanks{Chengyou Wang\\
	School of Mechanical, Electrical and Information Engineering, Shandong University, Weihai 264209, China \\
	Tel.: +86-631-5688338}
\thanks{e-mail: \href{mailto:wangchengyou@sdu.edu.cn?subject=CMFD}{wangchengyou@sdu.edu.cn}}
}

\markboth{}%
{}

\IEEEtitleabstractindextext{%
\begin{abstract}
The aim of this paper is to improve the accuracy of copy-move forgery detection (CMFD) in image forensics by proposing a novel scheme and the main contribution is evolving circular domains coverage (ECDC) algorithm. The proposed scheme integrates both block-based and keypoint-based forgery detection methods. Firstly, the speed-up robust feature (SURF) in log-polar space and the scale invariant feature transform (SIFT) are extracted from an entire image. Secondly, generalized 2 nearest neighbor (g2NN) is employed to get massive matched pairs. Then, random sample consensus (RANSAC) algorithm is employed to filter out mismatched pairs, thus allowing rough localization of counterfeit areas. To present these forgery areas more accurately, we propose the efficient and accurate ECDC algorithm to present them. This algorithm can find satisfactory threshold areas by extracting block features from jointly evolving circular domains, which are centered on matched pairs. Finally, morphological operation is applied to refine the detected forgery areas.  Experimental results indicate that the proposed CMFD scheme can achieve better detection performance under various attacks compared with other state-of-the-art CMFD schemes. 
\end{abstract}

}

\maketitle

\IEEEdisplaynontitleabstractindextext

\IEEEpeerreviewmaketitle

\ifclearsectionlook\cleardoublepage\fi \section{Introduction}
\label{sec:introduction}
\IEEEPARstart{W}{ith} 
With the development of computer and image processing software, digital image tampering becomes much easier; therefore, lots of digital images lack authenticity and integrity, which poses a threat to many critical fields. For example, it may lead to misdiagnosis when forged images are used in medical fields \cite{b1}, and forged newspaper photographs may mislead people and cause unnecessary social unrest \cite{b2}. Hence, the ability to credibly authenticate an image has become a major focus of image forensics and security.

The existing detection techniques fall into two main categories: active and passive. Active forensics techniques ensure the authenticity of digital images by verifying the integrity of authentication information, such as digital watermark \cite{b3,b4,b5} and digital signature \cite{b6,b7,b8}. These active methods have strong detection abilities and cannot be easily avoided, but their main defect is that the watermark must be inserted as a key into the image. Passive forensics techniques are used to verify the authenticity by analyzing the information and structure of the image, which overcomes the disadvantage of active forensics techniques.

There are two main forgeries that alter the contents of images: splicing and copy-move. The common splicing forgery method consists in copying and pasting a part of an image into another image, while the copy-move forgery method is a way to copy and paste a part of an image into the same image. In recent years, copy-move forgery has become one of the most popular subtopics in forgery detection \cite{b9}. To make copy-move tampered images more trustworthy, some processing methods are probably required, including rotation, scaling, downsampling, JPEG compression, and noise addition. Considering that image copy-move forgery detection (CMFD) is a challenging topic, this paper focuses on CMFD algorithms.

The general steps of CMFD are feature extraction, feature matching, and postprocessing. Based on different extracting features, CMFD is divided into block-based, keypoint-based, and fusion of these two methods. The last one has become more popular in recent years.

In this paper, we propose a CMFD scheme based on evolving circular domains coverage (ECDC), which combines block-based and keypoint-based methods. It extracts two different descriptors from an image, and then we match and filter those descriptors to obtain a rough localization. After that, we employ the proposed ECDC algorithm to cover forgery areas. The refined forgery areas are obtained by postprocessing ultimately. The two main contributions of this paper are listed below:

1)	we combine the speed-up robust feature (SURF) in log-polar space and the scale invariant feature transform (SIFT) as descriptors to depict a host image more accurately. This not only raises the precision of the proposed scheme under plain copy-move forgery evidently, but also improves its robustness to various geometric transformations and signal processing.

2)	we propose a novel algorithm, named ECDC, to present forgery areas exactly. By comparing the differences of block features in jointly evolving circular domains, this algorithm which is based on the pre-positioning of keypoints can greatly reduce computational complexity and improve its running efficiency. In addition, ECDC cannot only cover large-scale tampered areas completely, but also depict small areas accurately.

The rest of this paper is organized as follows: Section 2 briefly reviews the related work of CMFD; Section 3 displays the framework of the proposed CMFD scheme and then explains each step in detail; Section 4 shows the experimental results of CMFD and their analysis; finally, Section 5 gives the conclusion.

\ifclearsectionlook\cleardoublepage\fi \section{Related Work}
\label{sec:related work}
In this section, we review some classic and state-of-the-art CMFD schemes. Based on the difference of extracted features from the image, we divide this review into the following parts: block-based method, keypoint-based method, and fusion of the two methods.

\subsection{Block-Based Methods}
The block-based CMFD methods, in general, divide a host image into small, regular, and overlapped blocks. After extracting features of each subblock, the results are obtained by matching and postprocessing those features. Fridrich \emph{et al}. \cite{b10} proposed the CMFD algorithm, which is a milestone in the field of CMFD. They used quantified discrete cosine transform (DCT) coefficients as features. Then, a lexicographically ordered feature matrix reducing the range of feature matching, was used to detect similar regions \cite{b9}. Popescu and Farid \cite{b11} used principal components analysis (PCA) as features to detect tampered areas. Bayram \emph{et al}. \cite{b12} proposed Fourier-Mellin transform (FMT) to extract features. They applied counting bloom filters instead of Lexicographic sorting as a matching scheme which was more efficient. Wang \emph{et al}. \cite{b13,b14} used the Gaussian pyramid to reduce the dimensions of images. The former used the Hu-moments of blocks, and the latter employed the mean value of image pixels in circle blocks, which were divided into concentric circles. Ryu \emph{et al}. \cite{b15} proposed a method based on rotationally-invariant Zernike moments, which can detect forged regions even though they are rotated. Li \cite{b16} proposed an algorithm that matched polar cosine transform (PCT) with locality sensitive hashing (LSH), which required simpler calculations than Zernike moments. This algorithm excels at large-scale rotation. Similarly, polar sine transform (PST) and polar complex exponential transform (PCET) also belong to polar harmonic transform (PHT) \cite{b17}. Bravo and Nandi \cite{b40} used colour-dependent feature vectors to perform an efficient search in terms of memory usage. Cozzolino \emph{et al}. \cite{b18,b19} proposed a new matching method called PatchMatch, and a fast postprocessing procedure based on dense linear fitting. This method greatly reduces the computational complexity and it is robust to various types of distortions.

Overall, although applying Lexicographic sorting and reducing dimensions make block-based methods detection more efficient, it still has higher computational complexity than keypoint-based methods. In addition, when faced with large-scale scaling, the robustness of block-based methods, in general, is significantly reduced.

\subsection{Keypoint-Based Methods}
The keypoint-based CMFD methods usually extract features from an entire image, which is the main difference from block-based methods, and they effectively reduce computational complexity. Huang \emph{et al}. \cite{b20} proposed the best-bin-first nearest neighbor identification algorithm based on SIFT. Xu \emph{et al}. \cite{b21} proposed SURF to extract features with a faster speed compared with SIFT. Amerini \emph{et al}. \cite{b22} used generalized 2 nearest neighbor (g2NN) on SIFT descriptor to obtain qualified features. Then the random sample consensus (RANSAC) was used to remove mismatched points. Shivakumar and Baboo \cite{b23} proposed a CMFD scheme based on SURF and kd-tree for multidimensional data matching. In high-resolution image processing process, this method can detect different sizes of copied regions with a minimum number of false matches. To present tampered areas accurately, Pan and Lyu \cite{b24} utilized RANSAC to estimate the affine transformation matrix, and then they obtained correlation maps by calculating correlation coefficients to locate forged regions. Silva \emph{et al}. \cite{b44} proposed to separate forged points and the corresponding original ones into different clusters by clustering matched keypoints based on their locations and the final decision is based on a voting process. Park \emph{et al}. \cite{b43} utilized SIFT and the reduced local binary pattern (LBP) histogram to detect tampered areas.

However, \cite{b22,b23} only roughly marked the detected regions with connections on matched pairs. Furthermore, when tampering occurs in low-entropy or small-size areas, the detection results of many keypoint-based methods are unsatisfying due to the small number of keypoints.

\subsection{Fusion of Block-Based and Keypoint-Based Methods}
For better detection performance, combining the advantages of block-based and keypoint-based methods have currently become a trend. Some researchers proposed to segment the host image into non-overlapped and irregular blocks and then to match features extracted from those segmented regions \cite{b39,b25}. But their accuracy depends on the size of superpixels and detected results may have fuzzy boundaries. Zheng \emph{et al}. \cite{b26} classified the host image into textured and smooth regions in which, SIFT and Zernike features were respectively extracted and matched. However, this method cannot accurately distinguish between smooth and textured areas, especially when tampered regions are attacked by noise. Zandi \emph{et al}. \cite{b27} proposed a new interest point detector and used an effective filtering algorithm and an iteration algorithm to improve their performance. Although they can effectively detect tampered areas in low contrast areas, their detected results usually contain mismatches. Pun and Chung \cite{b28} proposed a two-stage localization for CMFD. The weber local descriptor (WLD) was extracted from each superpixel in their rough localization stage, and in their precise localization stage, discrete analytic Fourier-Mellin transform (DAFMT) of roughly located areas were extracted. Li and Zhou \cite{b42} proposed a hierarchical matching strategy to improve the keypoint matching problems and an iterative localization technique to localize the forged areas. Wang \emph{et al}. \cite{b29} classified irregular and non-overlapping image blocks into smooth and textured regions. They combined RANSAC algorithm with a filtering strategy to eliminate false matches. This method can detect a high-brightness smooth forgery. However, these methods achieve high detecting accuracy at the expense of low efficiency. 

In summary, the main problems faced by block-based CMFD methods are the inability to detect images with large-scale scaling and high computational complexity, while the main problem of keypoint-based CMFD methods is that there are fewer keypoints in low-entropy areas, which lead to incomplete coverage of tampered areas. Fusing block-based and keypoint-based methods reasonably can preserve their advantages and avoid certain shortcomings at the same time. Our scheme fairly integrates block-based and keypoint-based methods, which results in complete coverage of tampered areas and higher detection efficiency. The algorithm is described in more detail in Section 3.
\ifclearsectionlook\cleardoublepage\fi \section{Proposed Copy-Move Forgery Detection Scheme}
In this section, we explicate our CMFD scheme. The framework of the whole scheme is given in Fig. 1. Firstly, we extract both SIFT descriptor and log-polar SURF descriptor (LPSD) from an entire image. Secondly, g2NN is employed on each descriptor to obtain massive matched pairs. Then, we employ RANSAC to eliminate mismatched pairs. Finally, the ECDC algorithm is used to present entire forgery regions through those matched pairs. In the rest of this section, Section 3.1 explains the feature extraction algorithm combining SIFT and LPSD; Section 3.2 introduces the keypoints matching algorithm using g2NN; Section 3.3 describes the process of eliminating mismatched pairs by using RANSAC; Section 3.4 explains in detail how matched pairs are expanded to whole forgery regions by using ECDC algorithm.
\begin{figure}[h]
	\begin{center}
		\includegraphics[width=\linewidth]{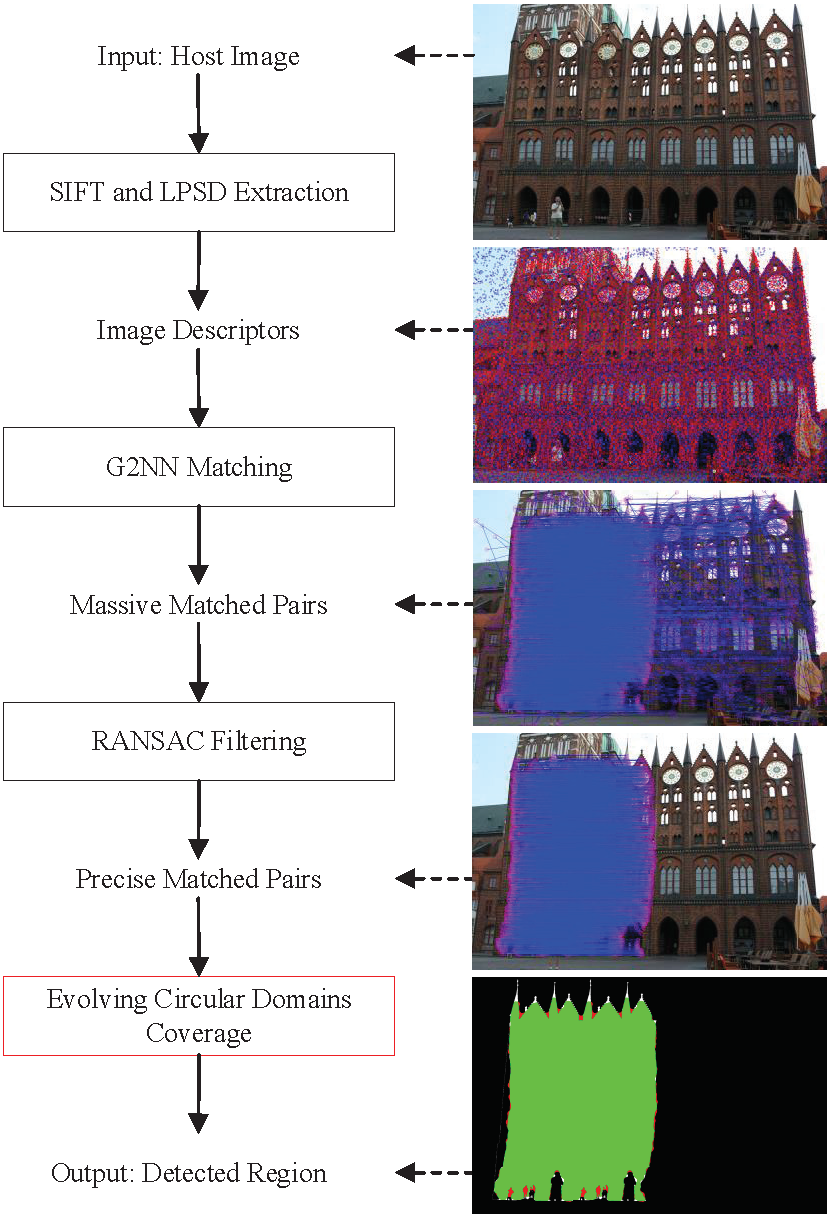}
	\end{center}
	\caption{Framework of the proposed copy-move forgery detection scheme.
		In the second image, SIFT is labeled with blue circles and LPSD is labeled
		with red dots.}
\end{figure}
\subsection{Feature Extraction Using Combination of SIFT and LPSD}
In this section, we explain how to extract keypoints as descriptors of the image. SIFT and SURF algorithms have been widely used in the field of computer vision in recent years. These keypoints are robust to various attacks, including rotation, scaling, downsampling, JPEG compression, and noise addition. As a result, SIFT and SURF are often used to extract keypoints in existing keypoint-based methods. In this paper, unlike in general keypoint-based methods, we combine SIFT and LPSD to depict images.

\subsubsection{SIFT}
Lowe \cite{b30} decomposed the SIFT algorithm into the following four steps: firstly, extrema in scale space were located with the computation searching over all scales and image locations; secondly, at each candidate location, keypoints were selected based on measures of their stability; then, based on local image gradient directions, one or more orientations were assigned to each keypoint location; at last, the local image gradients were measured at the selected scale in the region around keypoint to generate descriptors.

In general, the extreme points of a given image are detected at different scales in scale space, which is constructed by using the Gaussian pyramids with different Gaussian smoothing and resolution subsampling. These keypoints are extracted by applying difference of Gaussian (DoG), and a DoG image $D$ is denoted by \cite{b30}:

\begin{equation}
	\begin{split}
		D(x,y,\sigma)&=[G(x,y,k\sigma)-G(x,y,\sigma)]*I(x,y)\\
		&=L(x,y,k\sigma)-L(x,y,\sigma),
	\end{split}
\end{equation} where $L(x,y,k\sigma)$ is the convolution of the original image $I(x,y)$, with the Gaussian blur $G(x,y,\sigma)$ at scale space $k$.

To ensure rotation invariance, for each keypoint, the algorithm assigns a canonical orientation which can be determined by calculating the gradient in its neighborhood. Specifically, for an image sample $L(x,y,\sigma)$ at scale $\sigma$ , the gradient magnitude $m(x,y)$ and orientation $\theta(x,y)$ can be pre-calculated using pixel difference as follows \cite{b30}: 

\begin{equation}
	\begin{split}
		m(x,y) = &[[L(x + 1,y) - L(x - 1,y)]^2\\
		& + [L(x,y + 1) - L(x,y - 1)]^2 ]^\frac{1}{2},
	\end{split}
\end{equation}

\begin{equation}
	\theta (x,y) = {\tan ^{ - 1}}\frac{{L(x,y + 1) - L(x,y - 1)}}{{L(x + 1,y) - L(x - 1,y)}}.
\end{equation}

\subsubsection{SURF}
SURF proposed by Bay \emph{et al}. \cite{b31} is an improvement on SIFT, and being faster is its prominent characteristic. By using a Hessian matrix for optimization, SURF algorithm accelerates SIFT detection process without reducing the quality of the detected points. Then, box filters of different size are used to establish scale space and to convolute with the integral image. Given a point $\bm{x}=(x,y)$ in an image $\bm{I}$, the Hessian matrix $\bm{H}(\bm{x},\sigma) $ in $\bm{x}$ at scale $\sigma$ is represented as follows \cite{b31}:

\begin{equation}
	{\bm{{\rm H}}}({\bm{x}},\sigma ) = \left[ {\begin{array}{*{20}{c}}
			{{L_{xx}}({\bm{x}},\sigma )}&{{L_{xy}}({\bm{x}},\sigma )}\\
			{{L_{xy}}({\bm{x}},\sigma )}&{{L_{yy}}({\bm{x}},\sigma )}
	\end{array}} \right]
\end{equation} where $L_{xx}(\bm{x},\sigma)$ is the convolution result of the second order derivative of Gaussian filter with the image $\bm{I}$ in point $\bm{x}$, and similarly for $L_{xy}(\bm{x},\sigma)$ and $L_{yy}(\bm{x},\sigma)$.

Hessian matrix and non-maximum suppression are used to detect potential keypoints. While assigning one or more canonical orientations, the dominant orientation of the Gaussian weighted Harr wavelet responses can be detected by a sliding orientation window at every sample point within a circular neighborhood around the interest point.

\subsubsection{Combination of SIFT and LPSD}
Kaura and Dhavale \cite{b32} showed that the combination of SIFT and SURF would improve the detection performance of the keypoint-based method. In consideration of the lower detection accuracy of SURF, compared with SIFT \cite{b33}, we improve this accuracy by applying log-polar coordinates to it \cite{b34}. It can be seen in experiments that SURF in log-polar space, whose detection results are much more accurate than SIFT, succeeds well in detecting plain copy-move forgery, especially for detailed objects. Fig. 2(a1)–(a3) and Fig. 2(b1)–(b3) show SIFT and LPSD matched results for plain copy-move forgery, respectively. (The matching algorithm is explained in Section 3.2). We can observe that LPSD can obtain more matched pairs on small or detailed areas from Fig. 2(a3) and (b3). However, SIFT exhibits a surprising stability when forgery regions are attacked by noise or any other manipulations, as shown in Fig. 2(a4) and (b4). In these two figures, noise with standard deviation of 0.1 has been added to the copied fragments. In this case, LPSD hardly detects any right matched pairs while SIFT performs well. Thus, we decide to combine SIFT and LPSD to improve the instability of LPSD and the accuracy of SIFT.

\begin{figure}[h]
	\begin{center}
		\includegraphics[width=\linewidth]{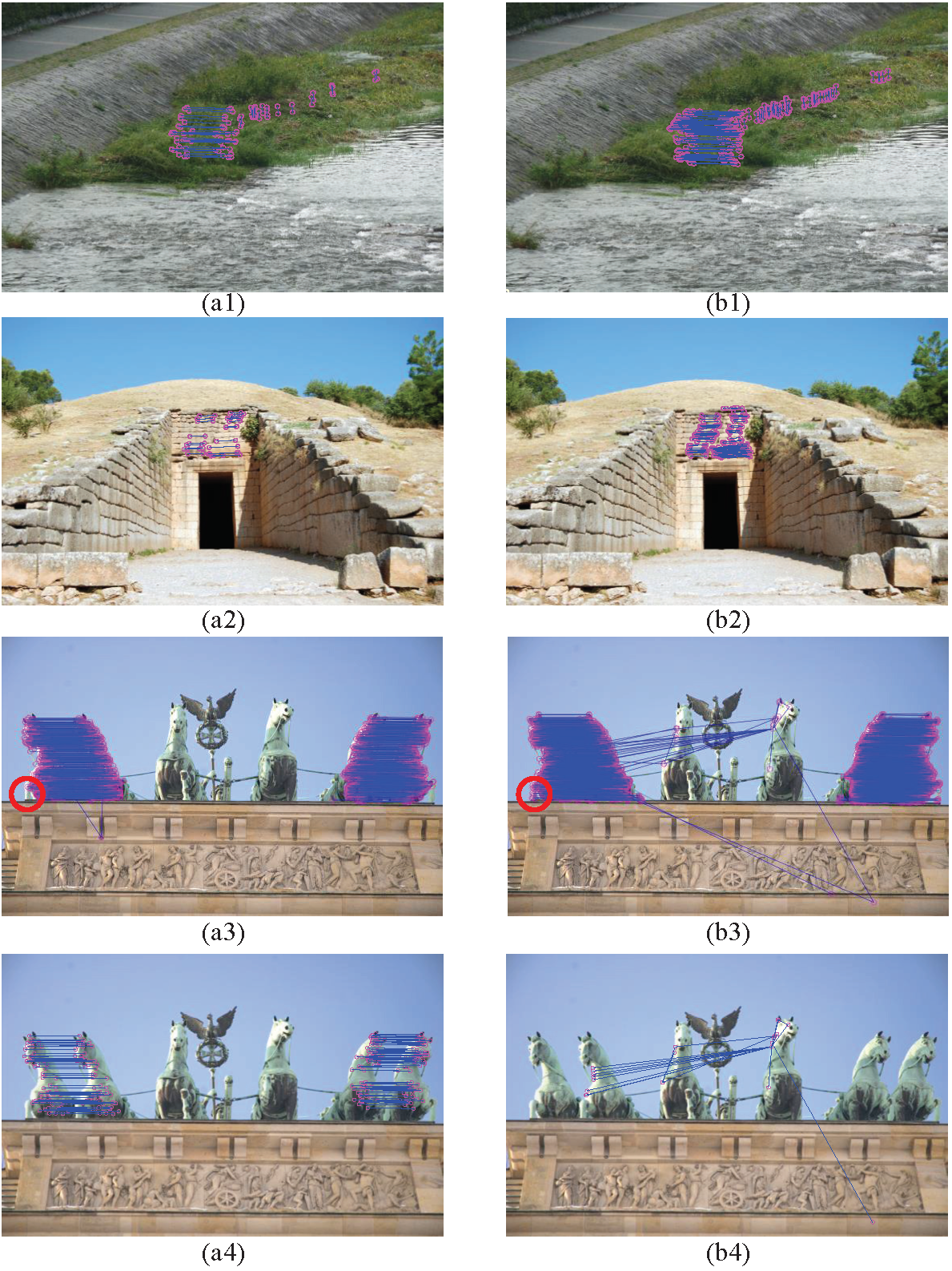}
	\end{center}
	\caption{Comparison of SIFT (left) and LPSD (right) detection results. First three
		rows: SIFT and LPSD detection results under plain copy-move forgery; fourth
		row: SIFT and LPSD detection results under local noise attack where the standard
		deviation is 0.1.}
\end{figure}

\subsection{Multiple Keypoints Matching}
\subsubsection{g2NN}
After feature extraction, two descriptor groups are obtained:
\begin{equation}
	{\bm{f}}^{\rm{SIFT}} = \{ {f_1}^{{\rm{SIFT}}},{f_2}^{{\rm{SIFT}}}, \cdot  \cdot  \cdot ,{f_n}^{{\rm{SIFT}}}\} ,
\end{equation}
\begin{equation}
	{\bm{f}}^{\rm{LPSD}} = \{ {f_1}^{{\rm{LPSD}}},{f_2}^{{\rm{LPSD}}}, \cdot  \cdot  \cdot ,{f_m}^{{\rm{LPSD}}}\} ,
\end{equation} where $\bm{f}^{\rm{SIFT}}$ is the $n$ dimensional SIFT descriptor vector and $\bm{f}^{\rm{\textsc{LPSD}}}$ is the $m$ dimensional LPSD descriptor vector. To find similar descriptors in the image, we need to match them to each other. Lowe \cite{b35} employed the distance ratio between the nearest neighbor and the second-nearest neighbor to compare it with a threshold $T$. Only if the ratio is less than $T$, the keypoints are matched. However, this matching process is unable to manage multiple keypoints matching. Since the same image areas may be cloned over and over in a tampered image, we employ g2NN algorithm \cite{b22} which can cope with multiple copies of the same descriptors. Specifically, taking SIFT as an example, we define a sorted distance vector ${{\bm{\chi }}_i}$ for $\bm{f}^{\rm{SIFT}}_i$ to represent the Euclidean distance between $\bm{f}^{\rm{SIFT}}_i$ and the other $(n-1)$ descriptors, i.e.,
\begin{equation}
	{{\bm{\chi }}_i} = \{ {d_{i,1}},{d_{i,2}}, \cdot  \cdot  \cdot ,{d_{i,n}}\} ,
\end{equation} where ${d_{i,j}}{\rm{ }}(i,j = 1,2, \cdots ,n;{\rm{ }}i\ne j)$ is the Euclidean distance between $\bm{f}^{\rm{SIFT}}_i$ and $\bm{f}^{\rm{SIFT}}_j$, i.e.,
\begin{equation}
	{d_{i,j}} = {\left\| {f_i^{{\rm{SIFT}}} - f_j^{{\rm{SIFT}}}} \right\|_2}.
\end{equation}
To facilitate the finding of an appropriate threshold, we measure the similarity between descriptors by using $d^2_{i,j}$(the Euclidean distance square). Thus, for all $\bm{f}^{\rm{SIFT}}$, an $n\times (n-1)$ matrix $\bm{\xi }$ will be generated:
\begin{equation}
	{\bm{\xi }} = \left[ {\begin{array}{*{20}{c}}
			{{\bm{\chi }}_{\rm{1}}^2}\\
			{{\bm{\chi }}_2^2}\\
			\vdots \\
			{{\bm{\chi }}_n^2}
	\end{array}} \right] = \left[ {\begin{array}{*{20}{c}}
			{d_{1,2}^2}&{d_{1,3}^2}& \cdots &{d_{1,n}^2}\\
			{d_{2,1}^2}&{d_{2,3}^2}& \cdots &{d_{2,n}^2}\\
			\vdots & \vdots & \ddots & \vdots \\
			{d_{n,1}^2}&{d_{n,2}^2}& \cdots &{d_{n,n - 1}^2}
	\end{array}} \right].
\end{equation}
We iterate 2 nearest neighbor (2NN) algorithm on every row of the distance matrix $\bm{\xi }$ to find multiple copies. Based on ${{\bm{\chi }}_i}$ as an example, the iteration will stop when
\begin{equation}
	{{d_{i,j}^2} \mathord{\left/
			{\vphantom {{d_{i,j}^2} {d_{i,j + 1}^2}}} \right.
			\kern-\nulldelimiterspace} {d_{i,j + 1}^2}} > T.
\end{equation}
If the iteration stops at $d^2_{i,k}$, each keypoint corresponding to the distance in $\{ d_{i,1}^2,d_{i,2}^2, \cdot  \cdot  \cdot ,d_{i,k}^2\} $ (where $k = 1,2 \cdots, n;{\rm{ }}k \ne i$) is considered as a match for the inspected keypoint.

\subsubsection{Threshold $T$}
Huang \emph{et al}. \cite{b20} analyzed that, if the ratio $T$ of the distance is reduced, then the number of matched keypoints will be reduced, but the matching accuracy will be improved. To test and verify this conclusion, we set different thresholds and observe the number of matched pairs and mismatched pairs of $\bm{f}^{\rm{SIFT}}$ and $\bm{f}^{\rm{LPSD}}$ under plain copy-move forgery. We use Figs. 3 and 4 to perceptibly and statistically describe the result. Fig. 3(a1)–(a4) and Fig. 3(b1)–(b4) show separately detected results of SIFT and LPSD, where thresholds range from 0.1 to 0.7 in steps of 0.2. To select an appropriate threshold, we randomly selected 100 images, including plain copy-move, rotation, scaling, noise, and other attacks, from the FAU dataset \cite{b9} for g2NN testing. Statistics data of SIFT and LPSD correct and wrong matches at different thresholds are respectively plotted as a line chart in Fig. 4(a) and (b).

\begin{figure}[h]
	\begin{center}
		\includegraphics[width=\linewidth]{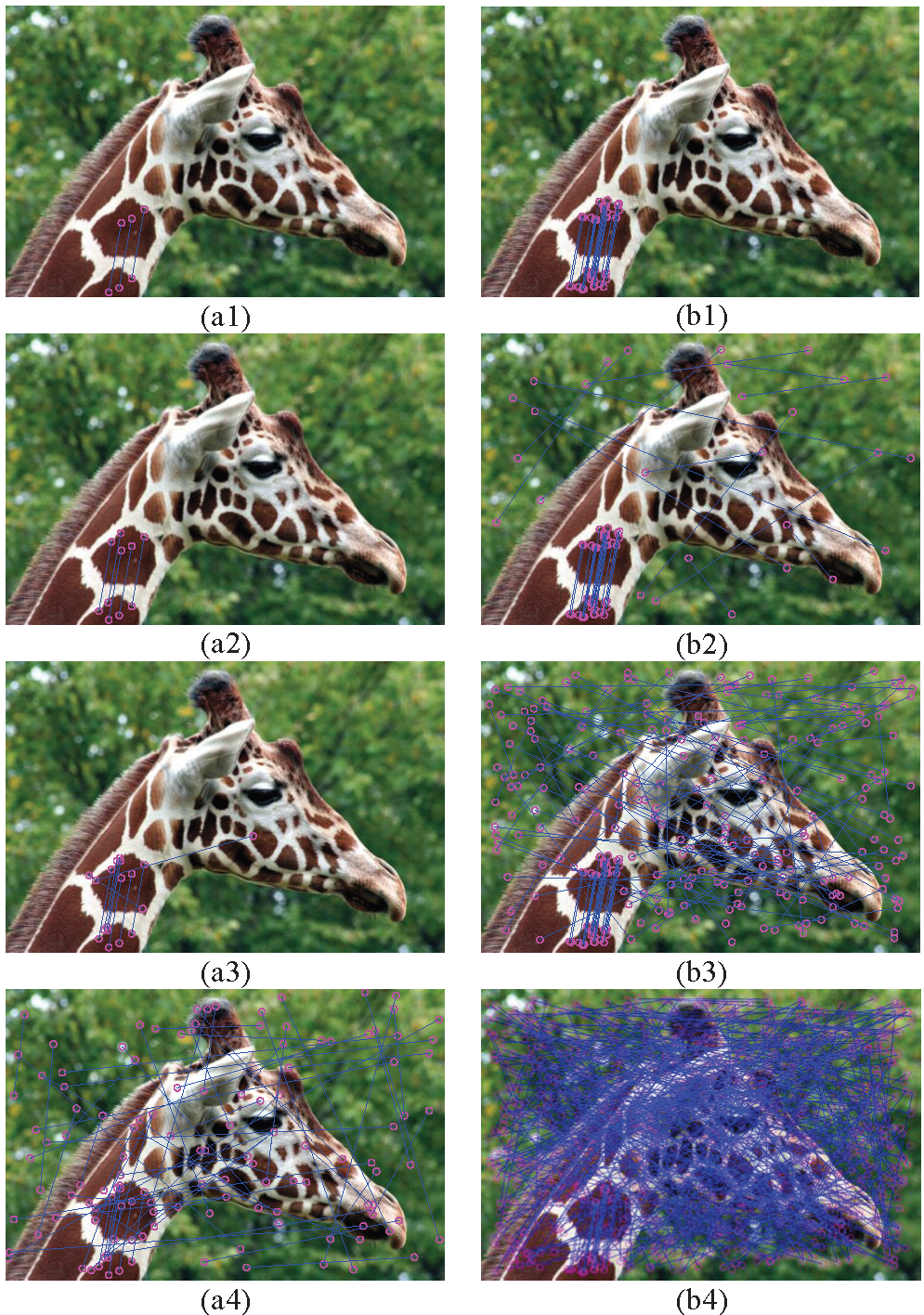}
	\end{center}
	\caption{Comparison of matches under different g2NN thresholds for SIFT (left)
		and LPSD (right) descriptors.}
\end{figure}

From Fig. 4, we can observe that with the increase of the threshold, correct matches tend to be constant, while incorrect matches increase rapidly. Thus, we come to two conclusions:

1)	A higher threshold will lead to more false matches, while a lower one may miss some correct matches. It is believed that appropriate threshold should not only obtain as many correct matches as possible, but also guarantee the number of incorrect matches within acceptable limits.

2)	Because LPSD has more mismatches at lower thresholds than SIFT descriptor, we set different g2NN thresholds $T_{\rm{SIFT}}$ and $T_{\rm{LPSD}}$ for them. The parameters used are presented in Section 4.1.

\begin{figure}[h]
	\begin{center}
		\includegraphics[width=\linewidth]{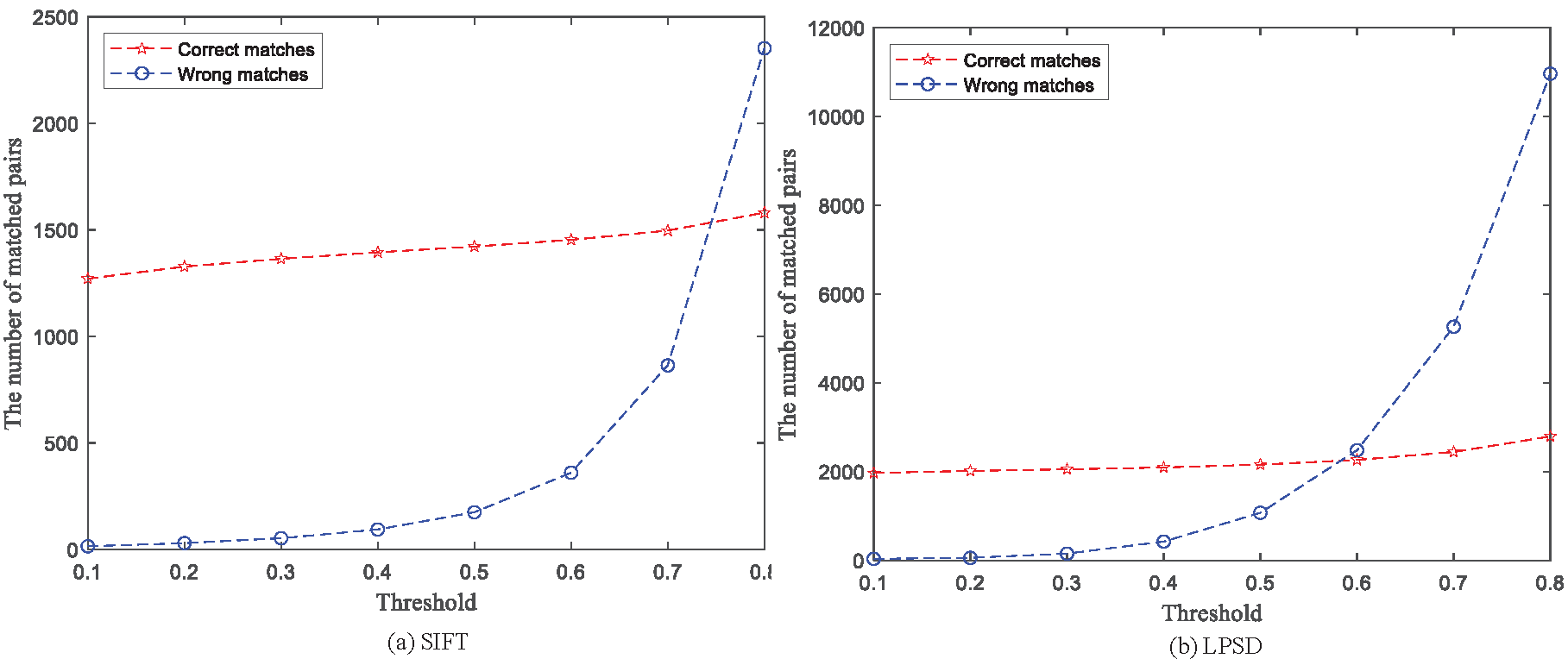}
	\end{center}
	\caption{The matching results using different g2NN thresholds with (a) SIFT and (b)
		LPSD. The correct numbers of matches are depicted in red dashed line with
		pentagon, while the wrong matches are depicted in dashed line with blue circle.}
\end{figure}

\subsection{Multiple Keypoints Matching}
After keypoints matching, we will get a large number of matched pairs. Due to the fact that adjacent keypoints have high similarity, we must remove the matched pairs when
\begin{equation}
	\sqrt {{{({x_a} - {x_b})}^2} + {{({y_a} - {y_b})}^2}}  < S,
\end{equation} where $(x_a,y_a)$ and $(x_b,y_b)$ indicate the coordinates of matched keypoints.

However, after that, many mismatched pairs still remain, which will seriously put an negative impact on covering or presenting forgery areas. Thus, we employ a widely used and robust algorithm named RANSAC \cite{b36} to eliminate them. RANSAC algorithm can estimate a model parameter precisely even when there are lots of mismatched pairs. It divides those pairs into inlier and outlier groups. To get enough matched pairs and, at the same time, to eliminate mismatched pairs with high similarity, our RANSAC algorithm is based on \cite{b29}.

We set the threshold $N$ and repeat RANSAC algorithm until the inlier groups points number is less than $N$. The higher $N$ is, the more mismatched pairs are eliminated. Meanwhile, those slight forgery or low-entropy regions are more likely to be overlooked. By contrast, a lower $N$ is better for detecting those regions. However, it can cause difficulty in eliminating mismatched pairs with high similarity. Therefore, we should get the right balance between the two contradictions.

\subsection{Forgery Areas Coverage Algorithm}
After postprocessing, we get a number of precisely matched pairs; however, these matched pairs can only cover tampered areas partially, which means that the original appearance of those areas cannot be fully revealed. Hence, accurately covering tampered areas is pivotal in CMFD.

In fact, the matching results of block-based methods and keypoint-based methods are essentially the position of two sets of pixels. More specifically, for the coverage of tampered areas, block-based methods require to compare many image block features centered on pixels. If the features of two blocks are sufficiently similar, their central pixels are recorded as a pair of matched points and their corresponding blocks are subsequently covered. Similarly, we consider that keypoint-based methods can also achieve the goal of covering tampered areas by comparing features, which are within a certain range and centered on pixel points. With the help of keypoint prepositioning, the algorithm complexity can be greatly reduced, thereby improving its detection speed. Thus, we propose a new algorithm to cover tampered areas, which is called ECDC.

\subsubsection{Selection of an appropriate feature}
Then, for a better coverage, we analyzed and discussed a variety of features. Christlein \emph{et al}. \cite{b9} listed most of the effective features, including four types: moment-based, dimensionality reduction-based, intensity-based, and frequency domain-based features. The DCT coefficients of the frequency domain-based features perform well against noise attacks. Wang \emph{et al}. \cite{b29}, through experiments, concluded that PCET moments perform better than other moment-based features under various geometric transformations. Therefore, we chose DCT coefficients and PCET moments for subsequent experiments.

\subsubsection{Block feature matching}
We extract block features from two separate circular domains centered on a matched pair. Then, we compare those features, and if they are similar enough, the corresponding circular domains will be covered. However, different features have different ways to measure their similarity. We usually employ the Euclidean distance to measure the resemblance of PCET moments because its dimension is constant. If the Euclidean distance between ${\bm{F}}_{\rm{1}}^{{\rm{PCET}}}$ and ${\bm{F}}_{\rm{2}}^{{\rm{PCET}}}$ is smaller than the predefined threshold $K_{\rm{PCET}}$, it will be considered as a matched pair, i.e.,
\begin{equation}
	{\left\| {{\bm{F}}_1^{{\rm{PCET}}} - {\bm{F}}_2^{{\rm{PCET}}}} \right\|_2} < {K_{{\rm{PCET}}}}
\end{equation}
Concerning the DCT coefficients, the dimension of the matrices depends on the size of sub image blocks. Consequently, large sub image blocks are stored in grand matrices, which is not conducive to computation. Thus, we use singular value decompositions (SVD) \cite{b37,b38} to decompose the extracted DCT coefficients matrices, i.e.,
\begin{equation}
	{{\bm{F}}^{{\rm{DCT}}}} = {\bm{U\Lambda }}{{\bm{V}}^{\rm{T}}},
\end{equation} where $\bm{U}$ and $\bm{V}$ are unitary matrices and $\bm{\Lambda}$ is a diagonal matrix whose entries are the singular values of $\bm{F}^{\rm{DCT}}$. Since $\bm{\Lambda}$ contains the basic information of $\bm{F}^{\rm{DCT}}$, and its maximum value includes most of the basic information of $\bm{F}^{\rm{DCT}}$, we choose the maximum value $\lambda$ of $\bm{\Lambda}$ to represent $\bm{F}^{\rm{DCT}}$ of a circular domain, i.e.,
\begin{equation}
	\lambda  = \max ({\bm{\Lambda }}).
\end{equation}
If the difference between $\lambda_1$ and $\lambda_2$ of two circular domains is less than the threshold $K_{\rm{DCT}}$, i.e.,
\begin{equation}
	\left| {{\lambda _1} - {\lambda _2}} \right| < {K_{{\rm{DCT}}}},
\end{equation} we determine that these two circular domains are tampered areas. We take 48 images from the FAU dataset \cite{b9}, crop their center into sub image blocks of $3\times 3$, $39\times 39$, and $75\times 75$ sizes, and attack them in various ways. Then, we calculate the mean value of $\lambda$ (denoted as $\bar \lambda$) in these three sets of sub image blocks, and list the results in Table 1. It shows that $\bar \lambda$ has only a slight difference under various attacks, which proves the feasibility of representing $\bm{F}^{\rm{DCT}}$ by $\lambda$ to depict sub image blocks.

\begin{table}[htbp]
	\caption{COMPARISON OF IMAGE BLOCKS $\bar\lambda$ UNDER DIFFERENT ATTACKS.}
	\centering
	{\setlength{\tabcolsep}{5.5mm}{\begin{tabular}{@{}llll@{}} \toprule
			Image Types & $\bar\lambda(3\times 3)$ & $\bar\lambda(39\times 39)$ & $\bar\lambda(75\times 75)$\\ \midrule
			Original & 328.87 & 3867.53 & 7454.32\\
			Rotation(15°) & 317.85 & 3866.56 & 7434.00\\
			scaling(98\%) & 328.85 & 3800.30 & 7314.76\\
			Noise(0.06) & 330.31 & 3891.77 & 7496.27\\ \bottomrule
	\end{tabular}}}
\end{table}

The selection of the aforementioned thresholds $K_{\rm{PCET}}$ and $K_{\rm{DCT}}$ has a great influence on the accuracy and robustness of our algorithm. If they decrease, the criteria get more stringent and the coverage is more precise; however, if the image is attacked by noise and geometric transformations, this algorithm would more easily miss or misjudge tampered areas. On the contrary, it is more robust. These thresholds can be determined through a large number of experiments and they depend on the image resolution and attack type of datasets. For FAU dataset \cite{b9}, to make it more robust to noise attacks, we set those thresholds as functions ${K_{{\rm{PCET}}}}({\sigma _{\rm{s}}})$ and ${K_{{\rm{DCT}}}}({\sigma _{\rm{s}}})$, where $\sigma _{\rm{s}}$ is the difference of variance between two circular domains. Finally, based on numerous experiments, we have established two empirical formulas for ${K_{{\rm{PCET}}}}({\sigma _{\rm{s}}})$ and ${K_{{\rm{DCT}}}}({\sigma _{\rm{s}}})$, which are piecewise functions:
\begin{equation}
	{K_{{\rm{PCET}}}}({\sigma _{\rm{s}}}) = \left\{ {\begin{array}{*{20}{l}}
			{1{\rm{,}}}&{{\sigma _{\rm{s}}} \le 0.1{\rm{,}}}\\
			{25{\rm{,}}}&{0.1 < {\sigma _{\rm{s}}} \le 1{\rm{,}}}\\
			{75{\rm{,}}}&{{\sigma _{\rm{s}}} > 1,}
	\end{array}} \right.
\end{equation}
\begin{equation}
	{K_{{\rm{DCT}}}}({\sigma _{\rm{s}}}) = \left\{ {\begin{array}{*{20}{l}}
			{25,}&{{\sigma _{\rm{s}}} \le 1{\rm{,}}}\\
			{{\rm{50,}}}&{1 < {\sigma _{\rm{s}}} \le 10{\rm{,}}}\\
			{{\rm{100,}}}&{{\sigma _{\rm{s}}} > 10.}
	\end{array}} \right.
\end{equation}

\subsubsection{Circular domains evolution}
Since tampered areas sizes are uncertain, they may not be covered ideally if only the features within a single radius are used as coverage basis. Therefore, we set the radius to an evolving vector in steps of $\tau$:
\begin{equation}
	{\bm{R}} = \{ {r_1},{r_2}, \cdot  \cdot  \cdot ,{r_m}\} ,
\end{equation} where ${r_1} < {r_2} <  \cdot  \cdot  \cdot  < {r_m}$. In this way, we can compare the features of matched pairs in an evolving radius range by looping.

The detail of ECDC is illustrated in Fig. 5, in which radii changing process for a keypoint of a matched pair is shown in closeup. For ease of interpretation, the rings in the closeup are labeled with different colors. In the first comparison, we compare the features in the red ring centered on one of the matched pair. When the threshold $K \in \{ {K_{{\rm{PCET}}}},{K_{{\rm{DCT}}}}\}$ is met, the radius is enlarged to the size of the blue ring and a second round of comparison is made. If the difference between the features in the blue ring is still less than $K$, the radius continues to be enlarged until it reaches its maximum or the difference no longer fulfills that condition. Then, the previous radius is recorded and the loop is broken. After traversing all matched pairs with the above algorithm, their coverage is finally completed.

\begin{figure}[h]
	\begin{center}
		\includegraphics[width=\linewidth]{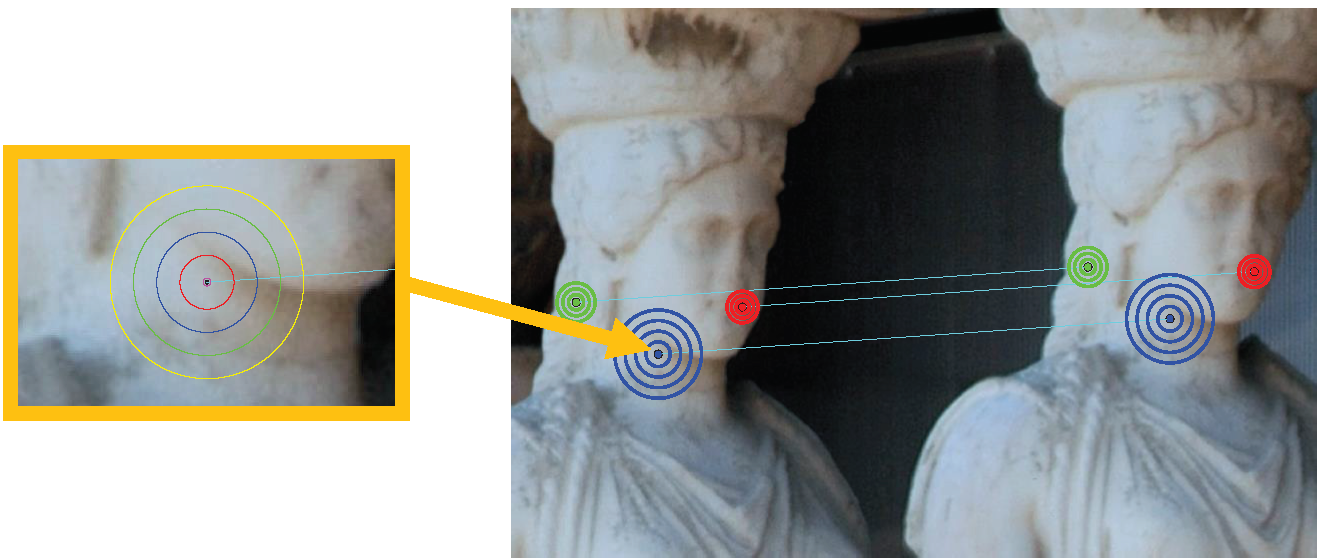}
	\end{center}
	\caption{Illustration of the detail of ECDC algorithm. Left: a locally enlarged
		radii changing process for a keypoint of a match. Right: the radii expansion
		process of three sets of matched pairs based on ECDC algorithm. Each matched
		pair is labeled with different colors.}
\end{figure}

The position of matched pairs is also of great importance on radii expansion. Three expansion results are presented in Fig. 5. The red and green pairs are near the edges of the tampered areas; thus, their rings' extension ends before the radius enlarged to its maximum $r_m$, which means that ECDC can accurately distinguish the edges. On the contrary, the blue pair is in the center of the tampered areas and, obviously, surrounded by it, so the expansion of the blue ring ends when the radius enlarges to its maximum $r_m$.

Fig. 6 presents the flowchart of ECDC algorithm, in which the middle image only partially shows the coverage of matched pairs. Furthermore, Fig. 5 represents the enlarged and detailed diagram of the step ‘Threshold Comparison Repetition’ of the loop in Fig. 6.
\begin{figure}[h]
	\begin{center}
		\includegraphics[width=\linewidth]{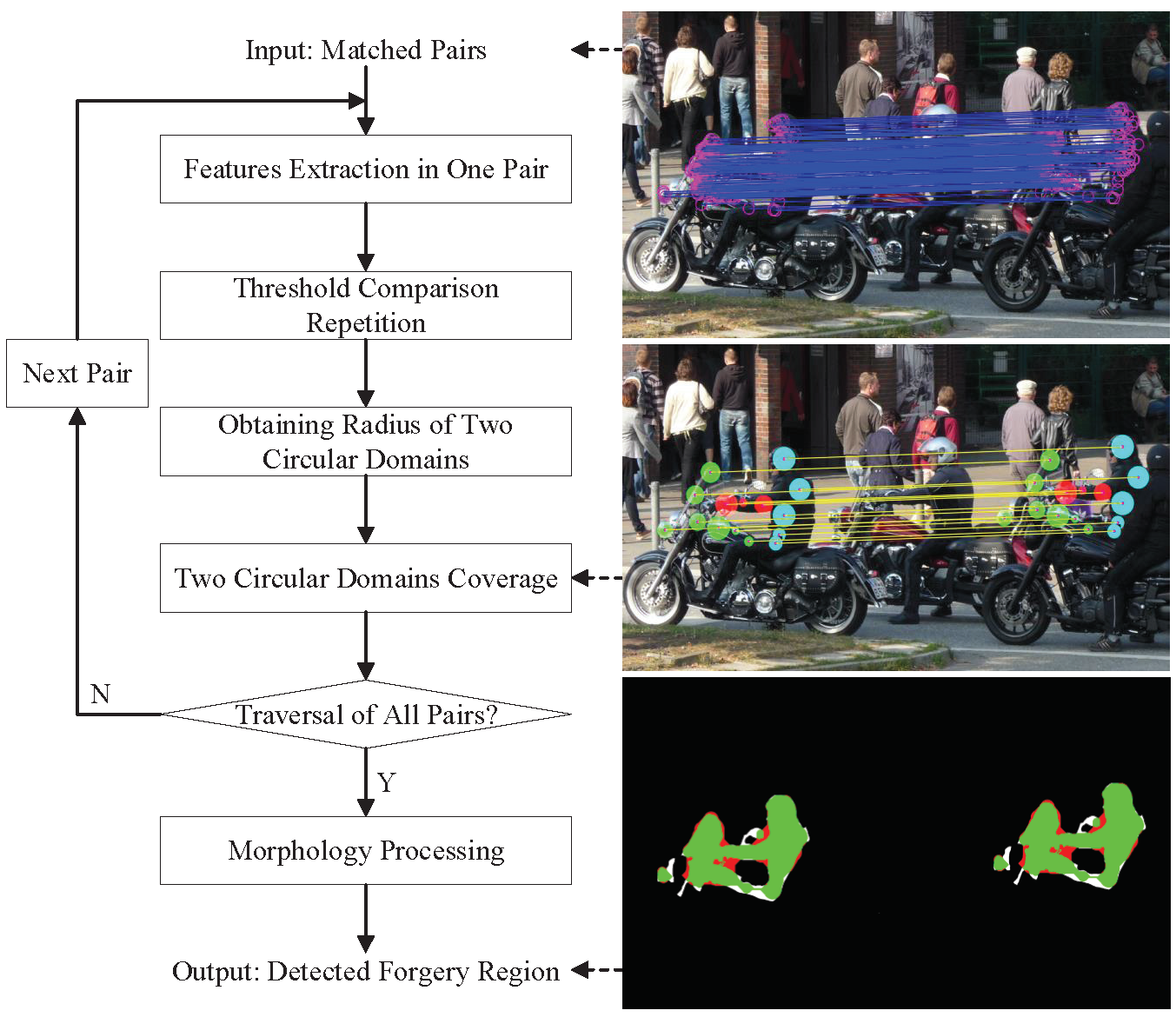}
	\end{center}
	\caption{Flowchart of ECDC algorithm.}
\end{figure}

\subsubsection{Morphological postprocessing}
Finally, depending on the image resolution, the disk size used for close operation varies. This step fills small holes and cracks in the merged areas while maintaining the overall outline of the areas as it is, which is advantageous to completely cover tampered areas.
\ifclearsectionlook\cleardoublepage\fi \section{Experimental Results and Analysis}
In this section, we conducted a series of experiments to compare validity and robustness between our scheme and other state-of-the-art schemes. Section 4.1 presents datasets we used, experimental setup, and parameters. Section 4.2 presents how we evaluated CMFD schemes. Section 4.3 presents the comparison between the proposed and other CMFD schemes at pixel level. Section 4.4 presents the comparison between the proposed and other CMFD schemes at image level.

\subsection{Image Datasets}
For a comprehensive comparison, three datasets, i.e., FAU \cite{b9}, GRIP \cite{b18}, and COVERAGE \cite{b41} are used to demonstrate the effectiveness of our scheme. FAU \cite{b9} dataset consists of 48 high-resolution images and it contains sub-datasets under various image attacks, including scaling, rotation, noise, downsampling, and JPEG compression. GRIP \cite{b18} only has plain copy-move images but some very smooth tampered areas, while COVERAGE \cite{b41} contains similar-but-genuine objects under a combination of different attacks. Hence, we chose FAU \cite{b9} to objectively evaluate CMFD schemes at pixel level, and GRIP \cite{b18}, COVERAGE \cite{b41} to evaluate them at image level. The detailed information of these three datasets is summarized in Table 2.

\begin{table*}[htbp]
	\caption{DETAILED INFORMATION OF POPULAR CMFD DATASETS WHICH USED IN OUR EXPRIMENTS INCLUDING FAU \cite{b9}, GRIP \cite{b18}, AND COVERAGE \cite{b41}.}
	\centering
	\renewcommand\arraystretch{1.2}
	{
	\setlength{\tabcolsep}{14mm}{\begin{tabular}{@{}llll@{}} \toprule
			Dataset & Average resolution & Number of images & Image format\\ \midrule
			FAU \cite{b9} & $1500\times1500$ & Authentic:48 Tampered:1968 & PNG\\
			GRIP \cite{b18} & $1024\times768$ & Authentic:80 Tampered:80 & PNG\\
			COVERAGE \cite{b41} & $400\times486$ & Authentic:110 Tampered:110 & TIF\\ \bottomrule
	\end{tabular}}}
\end{table*}

The experiments in this paper were performed in MATLAB 2019b on a 64-bit win10 PC with the Intel Core i7-8650 CPU model and 8 GB RAM.  Finally, we listed the parameters used in the proposed scheme in Table 3.

\begin{table}[ht]
	\caption{PARAMETERS SETTING IN THE PROPOSED SCHEME.}
	\centering
	\renewcommand\arraystretch{1.2}
	{
		\setlength{\tabcolsep}{7mm}{\begin{tabular}{@{}lll@{}} \toprule
			Parameter & Value & Meaning \\ \midrule
			$T_{\rm{SIFT}}$ & 0.6 & Threshold of SIFT in g2NN test \\
			$T_{\rm{LPSD}}$ & 0.1 &  Threshold of LPSD in g2NN test\\
			$S$ & 50 &  Threshold of Euclidean distance\\
			$r_1$ & 1.5 &  Minimum value of radii group\\ 
			$r_m$ & 37.5 &  Maximum value of radii group\\ 
			$\tau$ & 2 &  Step of radii group\\ \bottomrule
	\end{tabular}}}
\end{table}

\subsection{Evaluation Metrics}
Some state-of-the-art schemes uses True Positive Rate (TPR), False Positive Rate (FPR) and Accuracy (ACC) \cite{b42,b45} to evaluate their performance, while some choose precision $p$, recall $r$ \cite{b9,b25,b26} and $F_1$ score. To comprehensively evaluate CMFD methods, these two different metrics are used at two different levels.

At the image level, we focus on the practicality of our scheme to evaluate whether it can distinguish or not the difference between authentic images and forged images, as our original intention is to expose digital image forgery. In this case, metrics TPR, FPR and ACC are used. In CMFD schemes, the TPR $t$ indicates the percentage of correctly classified copy-move regions, while the FPR $f$ denotes that of incorrectly located cloned regions. They are defined as \cite{b42,b45}:

\begin{equation}
	\begin{split}
		t = \frac{N_{\rm{TP}}}{N_{\rm{TP}}+N_{\rm{FN}}}, f = \frac{N_{\rm{FP}}}{N_{\rm{TN}}+N_{\rm{FP}}},
	\end{split}
\end{equation} where $N_{\rm{TP}}$ is the number of correctly detected forged images, $N_{\rm{TN}}$ indicates the number of correctly detected authentic images, $N_{\rm{FP}}$ denotes the number of authentic images which have been erroneously detected as forged, and $N_{\rm{FN}}$ denotes the number of forged images which have not been detected. The accuracy of CMFD schemes $a$ denotes the performance of CMFD schemes based on TPR and FPR. It is defined as below \cite{b42,b45}:
\begin{equation}
	\begin{split}
		a = \frac{t+1-f}{2}, 
	\end{split}
\end{equation}
However, at the pixel level, we should not only pay attention if the proposed scheme can distinguish forged images and authentic images, but also cover detected forgery regions perfectly. In this case, precision $p$ and recall $r$ \cite{b9,b25,b26} are used to evaluate detection performance. Metrics $p$, $r$ and $F_1$ are defined as follows \cite{b9}:
\begin{equation}
	p = \frac{N_{\rm{TP}}}{N_{\rm{TP}} + N_{\rm{FP}}},
	r = \frac{N_{\rm{TP}}}{N_{\rm{TP}} + N_{\rm{FN}}},
\end{equation} where $N_{\rm{TP}}$ denotes the number of correctly detected forged pixels, $N_{\rm{FP}}$ denotes the number of pixels which has been erroneously detected as forged, and $N_{\rm{FN}}$ is the number of forged pixels which has not been detected. $p$ is used to describe the percentage of correctly detected pixels. A higher value of $p$ means there are less erroneous detections. $r$ describes whether the forgery areas are completely covered or not. A higher value of $r$ means the more complete the coverage of forgery areas is.

\begin{figure}[h]
	\begin{center}
		\includegraphics[scale=0.8]{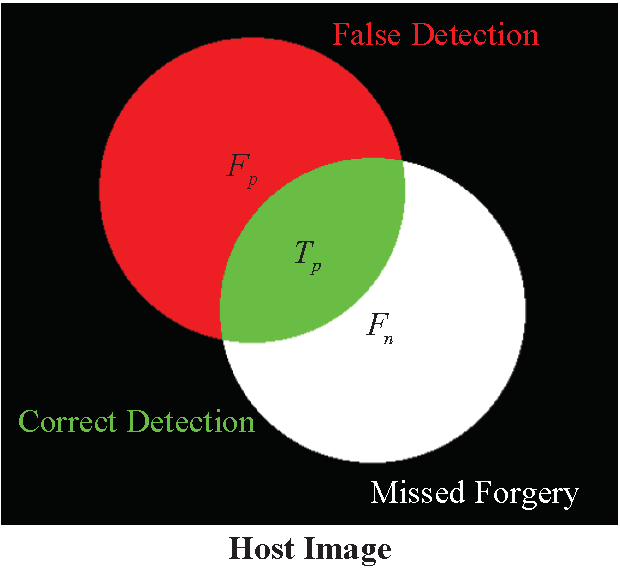}
	\end{center}
	\caption{A visualization of the relationship between $N_{\rm{TP}}$, $N_{\rm{FP}}$, and $N_{\rm{FN}}$ \cite{b28}.}
\end{figure}

By combining $p$ with $r$, the $F_1$ score is obtained \cite{b9}. The higher $F_1$ score gets, the better the performance is. 

\begin{equation}
	{F_1} = 2 \times \frac{{p \times r}}{{p + r}}.
\end{equation}

An intuitive illustration of the relationship between $N_{\rm{TP}}$, $N_{\rm{FP}}$, and $N_{\rm{FN}}$ is shown in Fig. 7. As the way of presenting in \cite{b27,b28} is clear, we employ the same way: green for correct detected areas, red for incorrect detected areas, and white for ground-truth areas, in which forged areas have not been detected.

\subsection{Detection Results Obtained on FAU at Pixel Level}
In this section, we mainly examined the ability of CMFD schemes to distinguish both authentic and forged images from FAU dataset\cite{b9} at pixel level. They should be able to show the forged areas in detail, which means they can perfectly display the particulars in the ideal situation. The performance of the proposed scheme is compared with that of various state-of-the-art CMFD methods, including block-based methods (e.g. \cite{b14,b15,b18}), keypoint-based methods (e.g. \cite{b20,b21,b22,b23,b24}) and fusion of both (e.g. \cite{b25,b26,b27}). 

\subsubsection{Plain CMFD}
We first evaluate their plain copy-move foregery detection performance. The detection results of the 48 images from the \emph{nul} sub-dataset are listed in Table 4, in descending $F_1$ order. The proposed scheme, while using DCT, achieves the optimal $F_1$, with $p=92.61\%$, $r=91.48\%$, and $p=91.56\%$. It has better CMFD performance at the pixel level compared with other algorithms. The highest $r$ is achieved when using PCET, because the coverage is more comprehensive; however, this leads to more coverage errors. Wang \emph{et al}. \cite{b30} achieved the highest $p$, which means they had the least number of detection errors. To sum up, our scheme reaches better results at the image level and the pixel level in the case of plain copy-move forgery. 

\begin{table}[htbp]
	\caption{DETECTION RESULTS UNDER PLAIN COPY-MOVE FORGERY AT THE PIXEL LEVEL IN DESCENDING $F_1$ ORDER}
	\centering
	{
		\setlength{\tabcolsep}{7mm}{\begin{tabular}{@{}llll@{}} \toprule
			Schemes & $p(\%)$ & $r(\%)$ & $F_1(\%)$\\ \midrule
			ECDC-DCT & 92.61 & 91.48 & 91.56\\
			Wang \cite{b14} & 98.69 & 85.44 & 90.92\\
			Ryu \cite{b15} & 95.07 & 87.72 & 90.29\\
			Cozzolino \cite{b18} & 92.98 & 88.98 & 90.19\\
			Pun \cite{b25} & 97.22 & 83.73 & 89.97\\
			Zheng \cite{b26} & 87.32 & 85.43 & 86.27\\
			ECDC-PCET & 81.84 & 93.31 & 86.09\\ 
			Zandi \cite{b27} & 83.65 & 79.53 & 79.66\\ 
			SURF \cite{b21,b23} & 68.13 & 76.43 & 69.54\\ 
			SIFT \cite{b20,b22,b24} & 60.80 & 71.48 & 63.10\\ \bottomrule
	\end{tabular}}}
\end{table}

\subsubsection{CMFD under Various Attacks}
As images are not only forged under plain copy-move manipulations, the robustness of different schemes should be tested especially when they are under various attacks. Therefore, sub-datasets of various types attacks are used, including scaling, rotation, noise, JPEG compression, downsampling, to present CMFD performance. Fig. 8 shows the detection results of our scheme under different attacks. The first and third columns represent forged images. The second and fourth columns are detection results. Fig. 8(a1) and (a3) show the plain copy-move forgery; Fig. 8(b1) and (b3) show the forged fragments scaled respectively by small and large scaling factors; Fig. 8(c1) and (c3) show the images under different rotation angles attacks; Fig. 8(d1) and (d3) show the local noise with two different standard deviations; Fig. 8(e1) and (e3) are the images under global noise attacks with two different standard deviations; Fig. 8(f1) and (f3) show the forged images attacked by JPEG compression with two different quality factors; Fig. 8(g1) and (g3) are two forged images downsampled by different downsampling factors. It can be seen from the results that our scheme also performs well on tampered images or forged fragments under various geometric transformations and signal processing.

\begin{figure*}[htbp]
	\begin{center}
		\includegraphics[width=\linewidth]{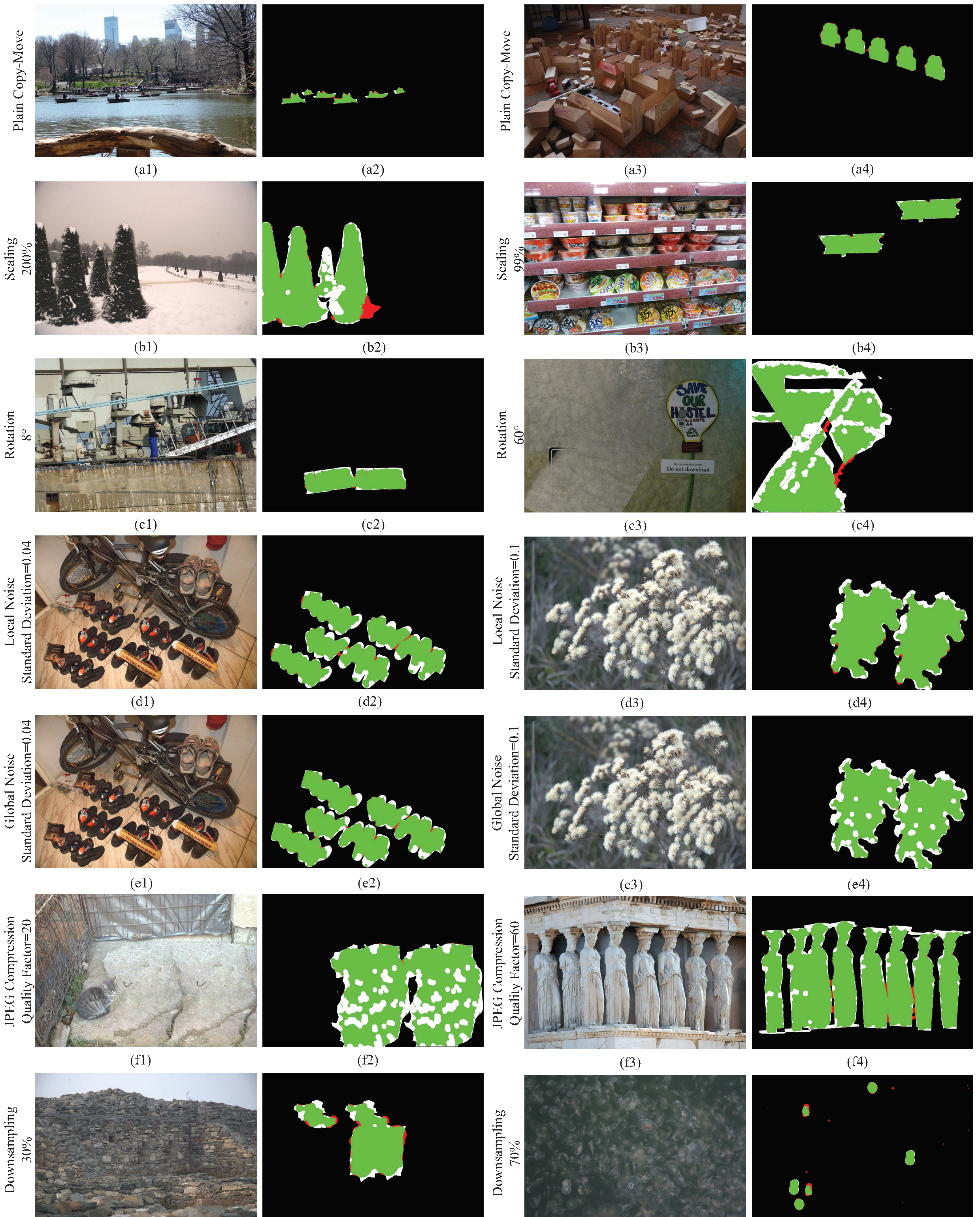}
	\end{center}
	\caption{Visualized CMFD results of the proposed scheme under various attacks where (a1), (a3), (b1), (b3), (c1), (c3), (d1), (d3), (f1), (f3), (g1), and (g3)
		respectively are \rm{center park, bricks, christmas hedge, supermarket, ship number, tapestry, sweets, white, lone cat, kore, stone ghost, and jellyfish chaos}. The first
		row shows the forgery images and the detection results under plain copy-move forgery. The second and third rows show the forgery images and the detection results
		under scaling and rotation. The fourth and fifth rows show the forgery images and the detection results under local noise and global noise, respectively. The sixth
		and last rows show the forgery images and the detection results under JPEG compression and downsampling, respectively. Green is the label of correctly detected
		regions and false areas are indicated in red. White color specifies the ground-truth areas, in which forged areas have not been detected.}
\end{figure*}

Figs. 9–11 show the $p$, $r$, and $F_1$ at the pixel level under (a) scaling, (b) rotation, (c) local noise, (d) global noise, (e) JPEG compression, and (f) downsampling attacks with different colors for different schemes’ results. Fig. 9 shows the $p$ results of the proposed scheme compared with the aforementioned schemes under different attacks. We can observe that the precision of the proposed scheme surpasses most of the others. Under small-scale rotation and scaling, the proposed scheme performs well, its precision with DCT is higher than that with PCET. In terms of large-scale rotation and scaling attacks, the results of the proposed scheme are superior to most of the others. The test results are displayed in Fig. 8(b2). Remarkably, the precision of our method running with DCT reaches more than 80\% at large-scale magnification. It also shows the highest results under the most severe global noise attacks. However, our scheme is affected by JPEG compression because of the extraction of many inoperative keypoints, especially when the quality factor is below 30. In this situation, as shown in Fig. 8(f2), our scheme can only maintain good performance to detect large forged areas, as it cannot filter out the invalid matches which are far more than the correct matches, when faced with small tampered areas. Of course, by making parameters of RANSAC and ECDC thresholds more stringent can reduce false coverage and the precision of JPEG compression with a low quality factor can be significantly improved; nevertheless, as the number of effective keypoints decreases, the precision of these results will considerably be reduced under local noise and global noise attacks. At this point, after strict filtering and ECDC, there will be only a few remaining matched pairs which do not have the ability to completely cover tampered areas. To sum up, it requires a compromise between performance under serious noise and under JPEG compression with a extremely low quality factor.

\begin{figure}[h]
	\begin{center}
		\includegraphics[width=\linewidth]{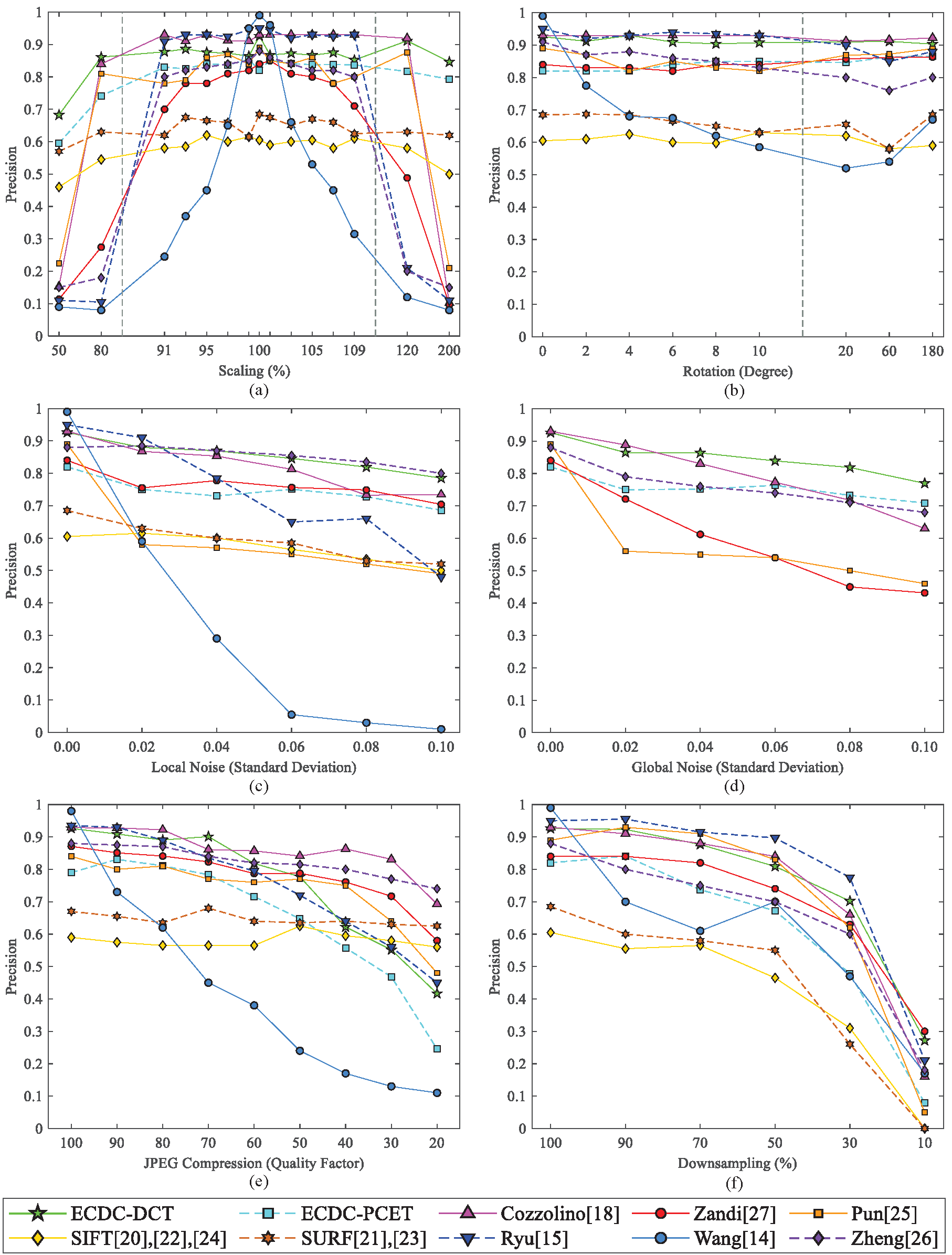}
	\end{center}
	\caption{Precision results at the pixel level: (a) Scaling; (b) Rotation; (c) Local
		noise; (d) Global noise; (e) JPEG compression; (f) Downsampling.}
\end{figure}

\begin{figure}[h]
	\begin{center}
		\includegraphics[width=\linewidth]{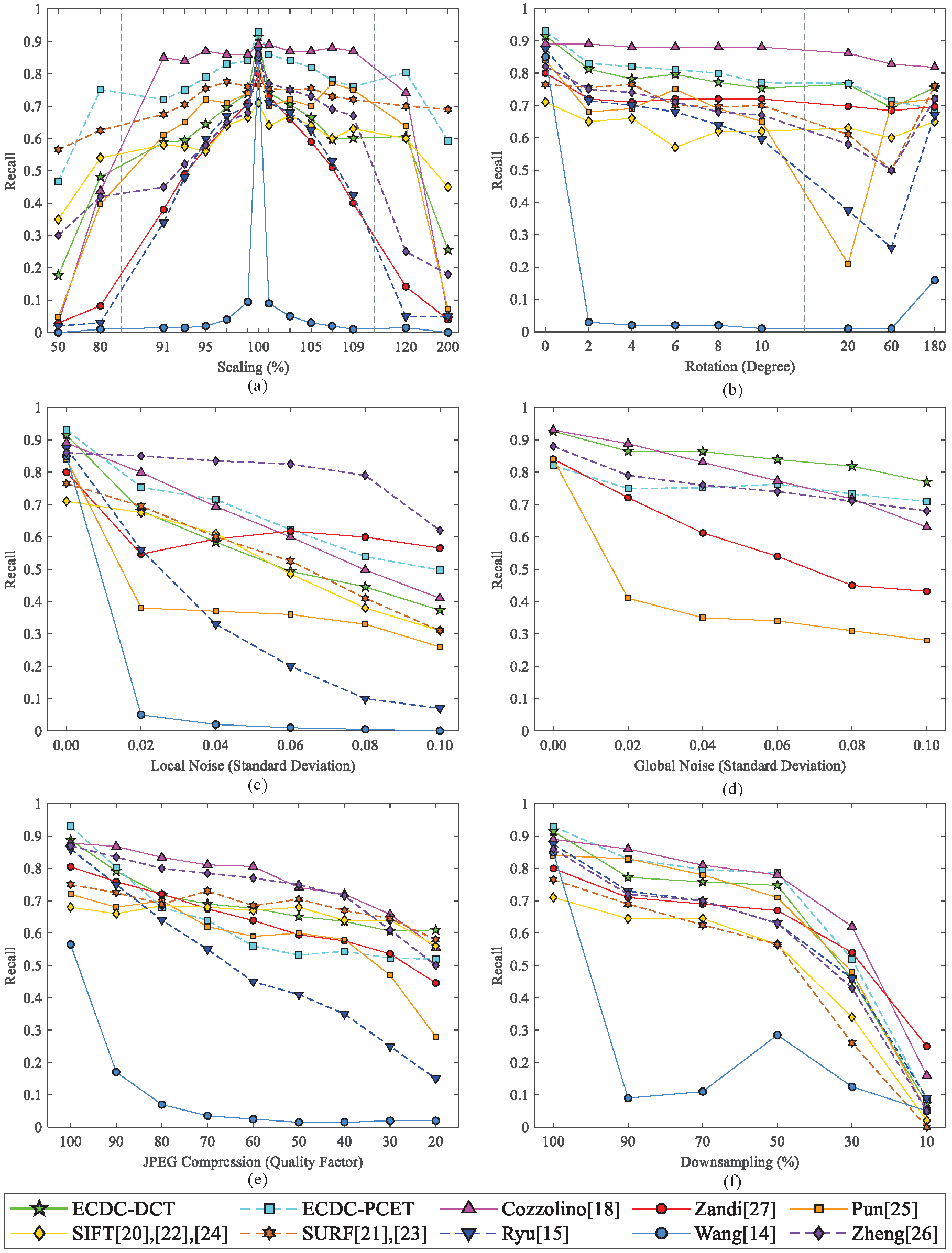}
	\end{center}
	\caption{Recall results at the pixel level: (a) Scaling; (b) Rotation; (c) Local
		noise; (d) Global noise; (e) JPEG compression; (f) Downsampling.}
\end{figure}

\begin{figure}[h]
	\begin{center}
		\includegraphics[width=\linewidth]{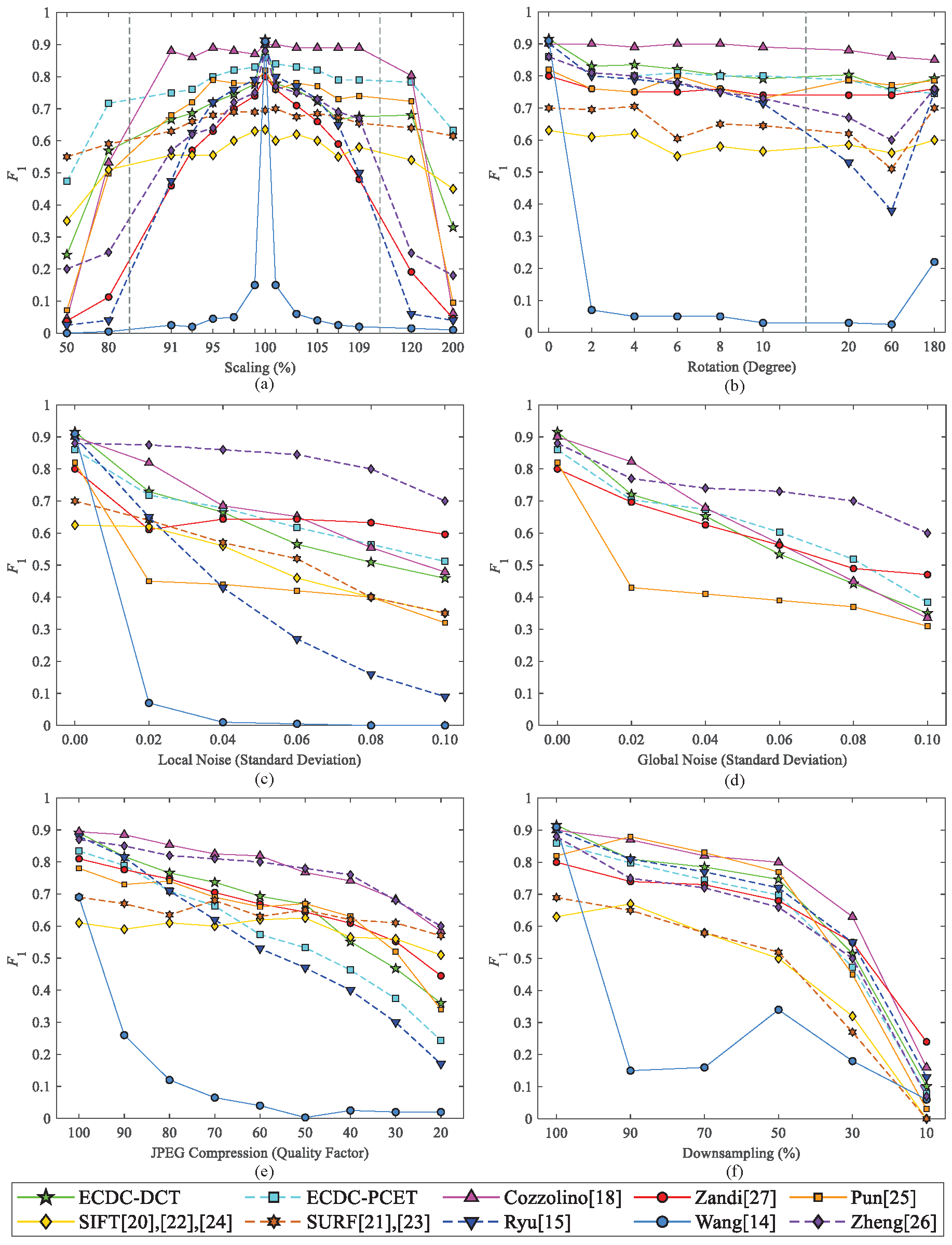}
	\end{center}
	\caption{$F_1$ scores at the pixel level: (a) Scaling; (b) Rotation; (c) Local noise;
		(d) Global noise; (e) JPEG compression; (f) Downsampling.}
\end{figure}

From Fig. 10, it also can be observed that the recall of our scheme with PCET is higher than that with DCT; therefore, we recommend using ECDC with PCET in vulnerable situations to cover forged areas more completely. Considering Figs. 9 and 10, we note that higher recall leads to lower precision, which means that the larger the coverage is, the lower the accuracy of detection may be. If tampered areas only have to be precisely indicated and they do not need to be presented perfectly, using ECDC with DCT would be a better option because it has higher detection precision with fewer mismatches. 

Fig. 11 depicts the comparison of all $F_1$ scores. We can intuitively conclude that ECDC is robust against all kinds of attacks whether it is in combination with DCT or PCET. Though the robustness of ECDC against some attacks is slightly inferior to the scheme \cite{b18}, it is exceptionally better in large-scale detection compared with most classic and state-of-the-art schemes tested.

\subsubsection{Running Time Comparison}
To comprehensively evaluate a CMFD scheme, we should pay attention to its running efficiency in addition to its effectiveness and reliability; thus, we also evaluate the processing efficiency of proposed scheme and others on FAU datasets. As the experimental platform of Christlein \emph{et al}. \cite{b9} is different from ours, we only compare the schemes avaiable and implemented on the same platform, and record the average running time of each scheme in Table 5, in an ascending order. It can be observed that our running time is relatively fast, and is above average compared with other solutions.

\begin{table}[htbp]
	\caption{RUNNING TIME OF THE PROPOSED SCHEME AND OTHER SCHEMES IN ASCENDING ORDER}
	\centering
	{
		\setlength{\tabcolsep}{12mm}{\begin{tabular}{@{}ll@{}} \toprule
			Schemes & Running Time (s)\\ \midrule
			Pun \cite{b25} & 128.45\\
			Cozzolino \cite{b18} & 149.03\\
			ECDC-DCT & 164.81\\
			Zandi \cite{b27} & 192.23\\
			ECDC-PCET & 275.87\\
			Zheng \cite{b26} & 554.36\\ \bottomrule
	\end{tabular}}}
\end{table}

\subsection{Detection Results Obtained on GRIP and COVERAGE at Image Level}
In this section, for a comprehensive and fair comparison, Other popular datasets, such as GRIP \cite{b18} and COVERAGE \cite{b41}, and metrics are used to evaluate state-of-the-art CMFD methods at image level. Fig. 12 illustrates serveral challenging forgery detection examples for these two datasets by using proposed method.

\begin{figure*}[h]
	\begin{center}
		\includegraphics[width=\linewidth]{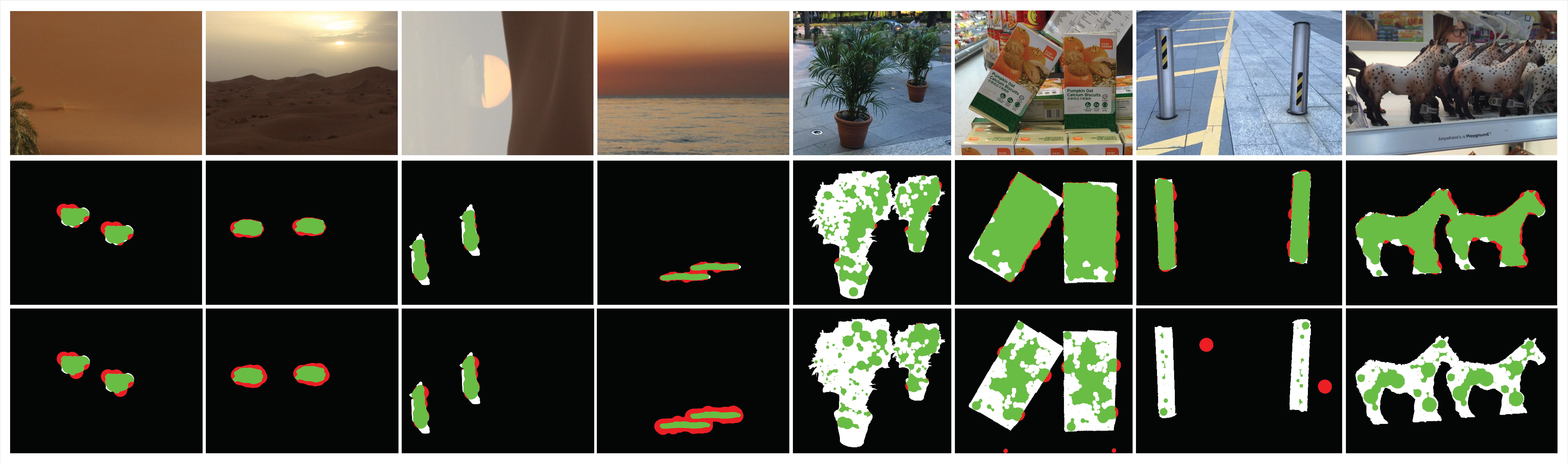}
	\end{center}
	\caption{Some challenging examples of copy-move forgery detection results on GRIP and COVERAGE. From top to bottom: forged images, results obtained by ECDC-DCT, and results obtained by ECDC-PCET. White color indicates the ground-truth; the correct detection pixels are marked in green, while those wrongly are in red.}
\end{figure*}

GRIP \cite{b18} contains some extremely smooth forged regions which is challenging for many keypoint-based methods, such as first three columns of Fig, 12. For comparison, keypoint-based methods \cite{b22,b44,b46}, block-based methods \cite{b40,b18} and fusion of both methods\cite{b39,b27,b42} are used. Table 6 presents the detection performance on this dataset, in descending ACC order. As shown in Table 6, both Li \cite{b42} and Bravo \cite{b40} exhibits the highest ACC of 100\%. The proposed CMFD algorithm using DCT and PCET achieves the second and third rank repectively, with an ACC of 98.75\% and 96.86\%. For this dataset, block-based methods \cite{b40,b18} and fusion of both methods\cite{b39,b27,b42} demonstrate generally better performance than keypoint-based methods \cite{b22,b44,b46}, due to challenging smooth tampered images.

\begin{table}[h]
	\caption{THE AVERAGE TPR $t$, FPR $f$, AND ACC $a$ OF DETECTION RESULTS ON GRIP AT THE PIXEL LEVEL IN DESCENDING $a$ ORDER}
	\centering
	{
		\setlength{\tabcolsep}{7mm}{\begin{tabular}{@{}llll@{}} \toprule
			Schemes & $t(\%)$ & $f(\%)$ & $a(\%)$\\ \midrule
			Li \cite{b42} & 100 & 0 & 100\\
			Bravo \cite{b40} & 100 & 0 & 100\\
			ECDC-DCT & 97.50 & 0 & 98.75\\
			ECDC-PCET & 95.00 & 1.25 & 96.86\\
			Cozzolino \cite{b18} & 98.75 & 8.75 & 95.00\\
			Chen \cite{b46} & 90.00 & 10.42 & 89.79\\
			Zandi \cite{b27} & 100 & 33.75 & 83.12\\ 
			Silva \cite{b44} & 100 & 38.75 & 80.63\\
			Amerini \cite{b22} & 70.00 & 20.00 & 75.00\\ 
			Li \cite{b39} & 83.75 & 35.00 & 74.38\\  \bottomrule
	\end{tabular}}}
\end{table}

Each image in COVERAGE \cite{b41} contains similar-but-genuine objects, resulting the fact that discrimination of forged from genuine objects is highly challenging. Moreover, many of their images are forged under a combination of image attacks. For comparison, keypoint-based methods \cite{b22,b43,b44}, block-based methods \cite{b40,b18} and fusion of both methods\cite{b39,b27,b42} are used. Table 7 shows the detection results on COVERAGE, in descending ACC order. It is obvious that all the algorithms perform poorly on this dataset. Silva \cite{b44} achieves the best TPR but the highest FPR high false positive rate, while Bravo \cite{b40} do not wrongly detected any authentic image as tampered one but it has the lowest TPR. Compared with the other algorithms, our method using DCT obtains the best ACC of 75.50\% and using PCET gets the third one.

\begin{table}[htbp]
	\caption{THE AVERAGE TPR $t$, FPR $f$, AND ACC $a$ OF DETECTION RESULTS ON COVERAGE AT THE PIXEL LEVEL IN DESCENDING $a$ ORDER}
	\centering
	{
		\setlength{\tabcolsep}{7mm}{\begin{tabular}{@{}llll@{}} \toprule
			Schemes & $t(\%)$ & $f(\%)$ & $a(\%)$\\ \midrule
			ECDC-DCT & 76.00 & 25.00 & 75.50\\
			Bravo \cite{b40} & 50.55 & 0 & 75.30\\
			ECDC-PCET & 69.00 & 22.00 & 73.50\\
			Li \cite{b42} & 80.22 & 41.76 & 69.23\\
			Cozzolino \cite{b18} & 59.34 & 21.98 & 68.68\\
			Park \cite{b43} & 78.00 & 43.00 & 67.50\\
			Amerini \cite{b22} & 85.71 & 54.95 & 65.38\\
			Li \cite{b39} & 87.91 & 63.74 & 62.09\\
			Silva \cite{b44} & 91.21 & 70.33 & 59.94\\
			Zandi \cite{b27} & 76.92 & 71.43 & 52.75\\ \bottomrule
	\end{tabular}}}
\end{table}

\ifclearsectionlook\cleardoublepage\fi \section{Conclusion}
Nowadays, the phenomenon of easy falsification of images has been a hot spot in the field of digital image forensics and information security. Copy-move forgery is one of the most common manipulations in image forgery. In this paper, we propose a new CMFD scheme based on ECDC. By using the combination of SIFT and LPSD extraction algorithms, we get both SIFT and LPSD descriptors of the entire image. In this way, those descriptors can get more detail features, while being more robust to various attacks. Then we use g2NN to gain a large number of matched pairs. After that, we use RANSAC to eliminate most of the mismatched pairs and gain more precise matched pairs; thus, forgery regions have been located roughly. Then, to get the accurate forgery regions, we propose ECDC algorithm, which can cover forgery regions according to the block features of evolving circular domains. Finally, we use morphological operations to improve the results of ECDC algorithm.

Nowadays, as the resolution of images gets higher, their size gets larger. Due to the complexity of the matching features step, block-based methods take too much time, and keypoint-based methods have difficulty in perfectly covering forgery regions. These two factors become our driving force to propose this scheme. In that way, we surmount the barriers caused by applying block features or keypoints alone.

We conduct a large number of experiments on the proposed scheme with satisfactory results to testify that it is an advanced scheme in CMFD field. Those results show both high effectiveness and efficiency, a notable increase in evaluation metrics and running speed. In comparison with other state-of-the-art CMFD schemes, the proposed scheme achieves more outstanding performance, especially under plain copy-move forgery.

In the future, we will strive to combine ECDC with more robust features, and enable it to cope with more various image attacks. Meanwhile, we will research a more flexible and reasonable way of fusing block-based and keypoint-based methods, so that it can get better performance and higher efficiency.

\iflongversion
\ifCLASSOPTIONcompsoc

\ifCLASSOPTIONcaptionsoff
  \newpage
\fi

\bibliographystyle{IEEEtran}

\begin{thebibliography}{100}
\providecommand{\url}[1]{#1}
\csname url@samestyle\endcsname
\providecommand{\newblock}{\relax}
\providecommand{\bibinfo}[2]{#2}
\providecommand{\BIBentrySTDinterwordspacing}{\spaceskip=0pt\relax}
\providecommand{\BIBentryALTinterwordstretchfactor}{4}
\providecommand{\BIBentryALTinterwordspacing}{\spaceskip=\fontdimen2\font plus
\BIBentryALTinterwordstretchfactor\fontdimen3\font minus
  \fontdimen4\font\relax}
\providecommand{\BIBforeignlanguage}[2]{{%
\expandafter\ifx\csname l@#1\endcsname\relax
\typeout{** WARNING: IEEEtran.bst: No hyphenation pattern has been}%
\typeout{** loaded for the language `#1'. Using the pattern for}%
\typeout{** the default language instead.}%
\else
\language=\csname l@#1\endcsname
\fi
#2}}
\providecommand{\BIBdecl}{\relax}
\BIBdecl

\bibitem{b1} S. Sharma and U. Ghanekar, “A rotationally invariant texture descriptor to detect copy move forgery in medical images,” in Proc. IEEE Int. Conf. Comput. Intell. Commun. Technol., Ghaziabad, India, 2015, pp. 795–798.

\bibitem{b2} Photo Tampering Throughout History. Accessed: Nov. 20, 2019. [Online]. Available: \underline{https://pth.izitru.com/2016\_02\_01.html}

\bibitem{b3} C. Wang, H. Zhang, and X. Zhou, “A self-recovery fragile image water-marking with variable watermark capacity,” Appl. Sci., vol. 8, no. 4, Apr. 2018, Art. no. 548.

\bibitem{b4} A. Zear, A. K. Singh, and P. Kumar, “A proposed secure multiple watermarking technique based on DWT, DCT and SVD for application in medicine,” Multimedia Tools Appl., vol. 77, no. 4, pp. 4863–4882, Feb. 2018.

\bibitem{b5} A. Shehab, M. Elhoseny, K. Muhammad, A. K. Sangaiah, P. Yang, H. Huang, and G. Hou, “Secure and robust fragile watermarking scheme for medical images,” IEEE Access, vol. 6, pp. 10269–10278, Feb. 2018.

\bibitem{b6} X. Wang, J. Xue, Z. Zheng, Z. Liu, and N. Li, “Image forensic signature for content authenticity analysis,” J. Visual Commun. Image Represent., vol. 23, no. 5, pp. 782–797, Jul. 2012.

\bibitem{b7} M. Okawa, “From BoVW to VLAD with KAZE features: Offline signature verification considering cognitive processes of forensic experts,” Pattern Recogn. Lett., vol. 113, pp. 75–82, Oct. 2018.

\bibitem{b8} M. Okawa, “Synergy of foreground–background images for feature extraction: Offline signature verification using Fisher vector with fused KAZE features,” Pattern Recogn., vol. 79, pp. 480–489, Jul. 2018.

\bibitem{b9} V. Christlein, C. Riess, J. Jordan, C. Riess, and E. Angelopoulou, “An evaluation of popular copy-move forgery detection approaches,” IEEE Trans. Inf. Forensics Secur., vol. 7, no. 6, pp. 1841–1854, Dec. 2012.

\bibitem{b10} A. J. Fridrich, B. D. Soukal, and A. J. Lukáš, “Detection of copy-move forgery in digital images,” in Proc. Digit. Forensic Res. Workshop, Cleveland, OH, USA, 2003, pp. 55–61.

\bibitem{b11} A. C. Popescu and H. Farid, “Exposing digital forgeries by detecting duplicated image regions,” Dept. Comput. Sci., Dartmouth College, Hanover, NH, USA, Tech. Rep. TR2004-515, 2004.

\bibitem{b12} S. Bayram, H. T. Sencar, and N. Memon, “An efficient and robust method for detecting copy-move forgery,” in Proc. IEEE Int. Conf. Acoust. Speech Signal Process., Taipei, Taiwan, 2009, pp. 1053–1056.

\bibitem{b13} J. Wang, G. Liu, Z. Zhang, Y. Dai, and Z. Wang, “Fast and robust forensics for image region-duplication forgery,” Acta Auto. Sin., vol. 35, no. 12, pp. 1488–1495, Dec. 2009.

\bibitem{b14} J. Wang, G. Liu, H. Li, Y. Dai, and Z. Wang, “Detection of image region duplication forgery using model with circle block,” in Proc. Int. Conf. Multimedia Inf. Networking Secur., Wuhan, China, 2009, pp. 25–29.

\bibitem{b15} S. J. Ryu, M. J. Lee, and H. K. Lee, “Detection of copy-rotate-move forgery using Zernike moments,” in Proc. Lect. Notes Comput. Sci., Alberta, Canada, 2010, pp. 51–56.

\bibitem{b16} Y. Li, “Image copy-move forgery detection based on polar cosine transform and approximate nearest neighbor searching,” Forensic Sci. Int., vol. 224, no. 1–3, pp. 59–67, Jan. 2013.

\bibitem{b17} P. T. Yap, X. Jiang, and A. C. Kot, “Two-dimensional polar harmonic transforms for invariant image representation,” IEEE Trans. Pattern Anal. Mach. Intell., vol. 32, no. 7, pp. 1259–1270, Jul. 2010.

\bibitem{b40} S. Bravo-Solorio and A. K. Nandi, “Exposing duplicated regions affected by reflection, rotation and scaling,” in Proc. International Conference on Acoustics, Speech and Signal Processing, 2011, pp. 1880–1883.

\bibitem{b18} D. Cozzolino, G. Poggi, and L. Verdoliva, “Efficient dense-field copy-move forgery detection,” IEEE Trans. Inf. Forensics Secur., vol. 10, no. 11, pp. 2284–2297, Nov. 2015.

\bibitem{b19} D. Cozzolino, G. Poggi, and L. Verdoliva, “Copy-move forgery detection based on PatchMatch,” in Proc. IEEE Int. Conf. Image Process., Paris, France, 2014, pp. 5312–5316.

\bibitem{b20} H. Huang, W. Gou, and Y. Zhang, “Detection of copy-move forgery in digital images using SIFT algorithm,” in Proc. Pacific-Asia Workshop Comput. Intel. Ind. Appl., Wuhan, China, vol. 2, 2008, pp. 272–276.

\bibitem{b21} B. Xu, J. Wang, G. Liu, and Y. Dai, “Image copy-move forgery detection based on SURF,” in Proc. Int. Conf. Multimedia Inf. Networking Secur., Nanjing, China, 2010, pp. 889–892.

\bibitem{b22} I. Amerini, L. Ballan, R. Caldelli, A. Del Bimbo, and G. Serra, “A SIFT-based forensic method for copy-move attack detection and transformation recovery,” IEEE Trans. Inf. Forensics Secur., vol. 6, no. 3, pp. 1099–1110, Sept. 2011.

\bibitem{b23} B. L. Shivakumar and S. S. Baboo, “Detection of region duplication forgery in digital images using SURF,” Int. J. Comput. Sci. Issues, vol. 8, no. 4–1, pp. 199–205, Jul. 2011.

\bibitem{b24} X. Pan and S. Lyu, “Region duplication detection using image feature matching,” IEEE Trans. Inf. Forensics Secur., vol. 5, no. 4, pp. 857–867, Dec. 2010.

\bibitem{b44} E. Silva, T. Carvalho, A. Ferreira, and A. Rocha, “Going deeper into copy-move forgery detection: Exploring image telltales via multi-scale analysis and voting processes,” Journal of Visual Communication and Image Representation, vol. 29, pp. 16–32, May 2015.

\bibitem{b43} J. Y. Park, T. A. Kang, Y. H. Moon, and I. K. Eom, “Copy-move forgery detection using scale invariant feature and reduced local binary pattern histogram,” Symmetry, vol. 12, no. 4, article no. 492, pp. 1–16, Apr. 2020.

\bibitem{b39} J. Li, X. Li, B. Yang, and X. Sun, “Segmentation-based image copy-move forgery detection scheme,” IEEE Trans. on Inf. Forensics and Security, vol. 10, no. 3, pp. 507–518, Mar. 2015.

\bibitem{b25} C. M. Pun, X. C. Yuan, and X. L. Bi, “Image forgery detection using adaptive oversegmentation and feature point matching,” IEEE Trans. Inf. Forensics Secur., vol. 10, no. 8, pp. 1705–1716, Aug. 2015.

\bibitem{b26} J. Zheng, Y. Liu, J. Ren, T. Zhu, Y. Yan, and H. Yang, “Fusion of block and keypoints based approaches for effective copy-move image forgery detection,” Multidimens. Syst. Signal Proc., vol. 27, no. 4, pp. 989–1005, Oct. 2016.

\bibitem{b27} M. Zandi, A. Mahmoudi-Aznaveh, and A. Talebpour, “Iterative copy-move forgery detection based on a new interest point detector,” IEEE Trans. Inf. Forensics Secur., vol. 11, no. 11, pp. 2499–2512, Nov. 2016.

\bibitem{b28} C. M. Pun and J. L. Chung, “A two-stage localization for copy-move forgery detection,” Inf. Sci., vol. 463–464, pp. 33–55, Oct. 2018.

\bibitem{b42} Y. Li and J. Zhou, “Fast and effective image copy-move forgery detection via hierarchical feature point matching,” IEEE Trans. Inf. Forensics Secur., vol. 14, no. 5, pp. 1307–1322, May 2019.

\bibitem{b29} C. Wang, Z. Zhang, Q. Li, and X. Zhou, “An image copy-move forgery detection method based on SURF and PCET,” IEEE Access, vol. 7, pp. 170032–170047, Dec. 2019.

\bibitem{b30} D. G. Lowe, “Object recognition from local scale-invariant features,” in Proc. IEEE Int. Conf. Comput. Vision, Kerkyra, Greece, 1999, pp. 1150–1157.

\bibitem{b31} H. Bay, T. Tuytelaars, and L. Van Gool, “SURF: Speeded up robust features,” in Proc. Lect. Notes Comput. Sci., Graz, Austria, 2006, pp. 404–417.

\bibitem{b32} W. C. N. Kaura and S. Dhavale, “Analysis of SIFT and SURF features for copy-move image forgery detection,” in Proc. Int. Conf. Innov. Inf., Embed. Commun. Syst., Tamil Nadu, India, 2017, pp. 1–4.

\bibitem{b33} R. C. Pandey, S. K. Singh, K. K. Shukla, and R. Agrawal, “Fast and robust passive copy-move forgery detection using SURF and SIFT image features,” in Proc. Int. Conf. Ind. Inf. Syst., Gwalior, India, 2014, Art. no. 7036519.

\bibitem{b34} T. Tao and Y. Zhang, “A scale-invariant keypoint detector in log-polar space,” in Proc. SPIE Int. Soc. Opt. Eng., Tokyo, Japan, vol. 10225, 2017, Art. no. 102250P.

\bibitem{b35} D. G. Lowe, “Distinctive image features from scale-invariant keypoints,” Int. J. Comput. Vis., vol. 60, no. 2, pp. 91–110, Nov. 2004.

\bibitem{b36} M. A. Fischler and R. C. Bolles, “Random sample consensus: A paradigm for model fitting with applications to image analysis and automated cartography,” Commun. ACM, vol. 24, no. 6, pp. 381–395, Jun. 1981.

\bibitem{b37} G. H. Golub and C. Reinsch, “Singular value decomposition and least squares solutions,” Numer. Math., vol. 14, no. 5, pp. 403–420, Apr. 1970.

\bibitem{b38} H. Andrews and C. Patterson, “Singular value decompositions and digital image processing,” IEEE Trans. Acoust. Speech Signal Process., vol. 24, no. 1, pp. 26–53, Feb. 1976.

\bibitem{b41} B. Wen, Y. Zhu, R. Subramanian, T. T. Ng, X. Shen, and S. Winkler, “COVERAGE—A novel database for copy-move forgery detection,” in Proc. IEEE International Conference on Image Processing, Phoenix, AZ, USA, 2016, pp. 161–165.

\bibitem{b45} A. Ferreira, S. C. Felipussi, C. Alfaro, P. Fonseca, J. E. Vargas-Muñoz, J. A. Dos Santos, and A. Rocha, “Behavior knowledge space-based fusion for copy-move forgery detection,” IEEE Transactions on Image Processing, vol. 25, no. 10, pp. 4729–4742, Oct. 2016.

\bibitem{b46} H. Chen, X. Yang, and Y. Lyu, “Copy-move forgery detection based on keypoint clustering and similar neighborhood search algorithm,” IEEE Access, vol. 8, pp. 36863–36875, 2020.

\end{thebibliography}

\end{document}